\documentclass[10pt,twocolumn,letterpaper]{article}

\usepackage{cvpr}              
\usepackage{subcaption}
\usepackage{graphicx}

\usepackage[dvipsnames]{xcolor}

\definecolor{cvprblue}{rgb}{0.21,0.49,0.74}
\usepackage[pagebackref,breaklinks,colorlinks,citecolor=cvprblue]{hyperref}
\newcommand{\tablestyle}[2]{\setlength{\tabcolsep}{#1}\renewcommand{\arraystretch}{#2}\centering\small}

\usepackage{tabulary,multirow,xspace}
\usepackage{subcaption}
\usepackage{caption}
\usepackage{float}
\usepackage{wrapfig} 
\usepackage[linesnumbered,ruled,vlined]{algorithm2e}
\usepackage{makecell}
\usepackage{pifont}
\newcommand{\cmark}{\ding{51}}%
\newcommand{\xmark}{\ding{55}}%
\usepackage{tcolorbox}
\usepackage{subcaption}
\usepackage{graphicx}
\usepackage{tcolorbox}

\usepackage[T1]{fontenc}

\newcommand{\var}{\texttt}
\newcommand{\VarSty}[1]{\textnormal{\ttfamily\color{blue!90!black}#1}\unskip}

\definecolor{plotblue}{rgb}{0.235, 0.471, 0.847}
\definecolor{plotred}{rgb}{0.796, 0.075, 0.149}
\definecolor{plotbrown}{rgb}{0.523, 0.223, 0}
\definecolor{plotpurple}{rgb}{0.8, 0.6, 1}
\definecolor{plotsamplepurple}{rgb}{0.8, 0.6, 1}
\definecolor{plotsampleblue}{rgb}{0.6, 0.8, 1}
\definecolor{plotsamplegreen}{rgb}{0.2, 0.784, 0.2}
\definecolor{plotsamplered}{rgb}{0.6, 0, 0}
\definecolor{plotblue}{rgb}{0.235, 0.471, 0.847}
\definecolor{plotred}{rgb}{0.796, 0.075, 0.149}
\definecolor{plotpurple}{rgb}{0.8, 0.6, 1}
\definecolor{plotsamplepurple}{rgb}{0.8, 0.6, 1}
\definecolor{plotsampleblue}{rgb}{0.6, 0.8, 1}
\definecolor{plotsamplegreen}{rgb}{0.2, 0.784, 0.2}
\definecolor{plotsamplered}{rgb}{0.6, 0, 0}
\definecolor{cpurple}{HTML}{8330a0} 
\definecolor{plum}{HTML}{9c277e} 

\DeclareSymbolFont{extraup}{U}{zavm}{m}{n}
\DeclareMathSymbol{\vardiamond}{\mathalpha}{extraup}{87}
\newcommand{\corresponding}{\textcolor{plum}{$^{\spadesuit}$}}
\newcommand{\projlead}{\textcolor{cpurple}{$^{\vardiamond}$}}
\newcommand\blfootnote[1]{%
  \begingroup
  \renewcommand\thefootnote{}\footnote{#1}%
  \addtocounter{footnote}{-1}%
  \endgroup
}

\def\paperID{} 
\def\confName{3DV\xspace}
\def\confYear{2025\xspace}

\title{Reason3D: Searching and Reasoning 3D Segmentation via\\ Large Language Model}

\author{Kuan-Chih Huang$^1$ \quad Xiangtai Li$^2$\projlead \quad Lu Qi$^1$\corresponding \quad Shuicheng Yan$^{2,3}$ \quad Ming-Hsuan Yang$^{1}$ \vspace{0.2cm}\\
$^1$University of California, Merced \quad $^2$Skywork AI, Singapore \quad $^3$Nanyang Technological University \vspace{0.2cm}\\
\url{https://reason3d.github.io/}
}

\begin{document}

\twocolumn[{
  \renewcommand\twocolumn[1][]{#1}
  \maketitle  
  \begin{center}
    \vspace{-8mm}
  \includegraphics[width=0.81\linewidth]{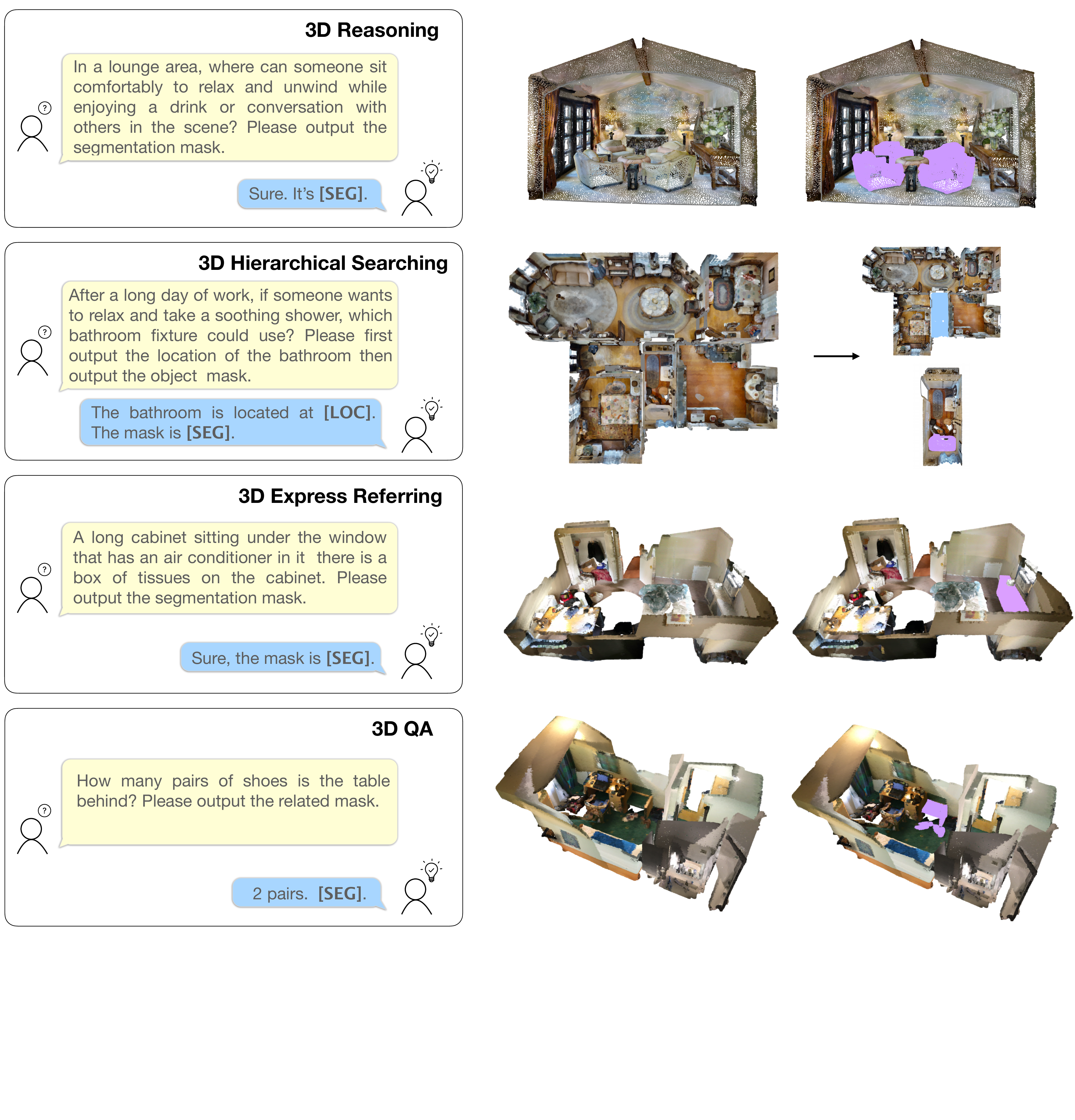}
  \vspace{-26mm}
\captionof{figure}{
\textbf{Overview.}
We propose \textbf{Reason3D}, a novel LLM-based 3D point cloud searching and reasoning framework that can output dense segmentation masks based on textural descriptions. 
Our Reason3D can handle four tasks involving 1) 3D Reasoning, 2) 3D Hierarchical Searching, 3) 3D Express Referring, and 4) 3D QA with responding dense segmentation masks.
}
  \label{fig:teaser}
  \end{center}
}]

\blfootnote{{\projlead Project lead, \corresponding Corresponding author}}

\begin{abstract}
Recent advancements in multimodal large language models (LLMs) have demonstrated significant potential across various domains, particularly in concept reasoning. 
However, their applications in understanding 3D environments remain limited, primarily offering textual or numerical outputs without generating dense, informative segmentation masks.
This paper introduces Reason3D, a novel LLM designed for comprehensive 3D understanding. Reason3D processes point cloud data and text prompts to produce textual responses and segmentation masks, enabling advanced tasks such as 3D reasoning segmentation, hierarchical searching, express referring, and question answering with detailed mask outputs.  
We propose a hierarchical mask decoder that employs a coarse-to-fine approach to segment objects within expansive scenes. 
It begins with a coarse location estimation, followed by object mask estimation, using two unique tokens predicted by LLMs based on the textual query.
Experimental results on large-scale ScanNet and Matterport3D datasets validate the effectiveness of our Reason3D across various tasks.
\end{abstract}    
\section{Introduction}
\label{sec:intro}
Recently, large language models (LLMs)~\cite{Achiam2023GPT4TR, llama, llama2} have significantly advanced their capabilities in sophisticated reasoning within natural language processing.
Building on these advancements, a new class of models known as Multimodal Large Language Models (MLLMs)~\cite{dai2023instructblip, li2023blip2, awadalla2023openflamingo, liu2023llava, zhang2023multicot, hanoona2023GLaMM,ye2023mplugowl,zhan2024anygpt,qi2024generalizable,liu2023llava,awadalla2023openflamingo,ren2023pixellm} has emerged, thereby enhancing LLMs' ability to interpret and understand visual inputs.

To advance the capabilities of Multimodal Large Language Models (MLLMs) in complex 3D environments, several studies have made significant strides by using point clouds as input tokens. 
Some research efforts~\cite{xu2023pointllm, GPT4Point, qi2024shapellm, tang2024minigpt_3d} have primarily focused on 3D object-level understanding. In addition, 3D-LLM~\cite{3dllm} aggregates multi-view features to enrich 3D feature comprehension and employs an LLM for subsequent 3D reasoning. LL3DA~\cite{chen2023ll3da} directly encodes 3D point clouds for scene representation, facilitating human interaction to enhance understanding.

These methods integrate large language models (LLMs) with point cloud inputs to enhance 3D reasoning capabilities; however, they face certain limitations. First, their outputs are restricted to textual or numerical forms, which are insufficient for predicting dense data types such as segmentation masks. Second, these models struggle to locate or identify objects in 3D scenes based on complex or abstract concepts, as they primarily rely on spatial relationships within the scenes.

To enable 3D-based LLM models to produce segmentation masks, we can direct the LLM to generate a \texttt{[SEG]} token. The embedding of this token is then utilized to guide the decoder in learning to predict 3D segmentation masks, similar to approaches used in 2D reasoning models~\cite{lai2023lisa}. 
However, unlike structured image data, this straightforward adaptation may encounter difficulties due to the inherent sparsity and unstructured nature of point clouds.
These challenges are particularly pronounced when segmenting small objects within large-scale 3D scenes.

%
This paper introduces Reason3D, a framework that enables reasoning and searching within 3D scenes using large language models with only point cloud inputs.   
Unlike other methods that generate only textual and numerical outputs, Reason3D also produces 3D segmentation masks from textual queries.
We first group point features into superpoints to reduce complexity and then utilize a transformer to align point features with textual instructions. These aligned features, along with query tokens, serve as the input for a pre-trained LLM.
%
%
To address the challenges of segmenting small objects within extensive point clouds, such as searching for a ball in a large house, we develop a hierarchical mask decoder that employs a coarse-to-fine strategy. 
This strategy begins by instructing the LLM to output the \texttt{[LOC]} and \texttt{[SEG]} token embeddings. 
The \texttt{[LOC]} token guides the learning of a region mask to identify likely object-containing areas. This region mask then serves as a prior, along with the segmentation token \texttt{[SEG]}, for generating a precise object mask, facilitating effective localization in complex 3D environments.

Figure~\ref{fig:teaser} illustrates Reason3D's capability to handle diverse tasks, such as reasoning, searching, referring, and question answering.
To validate the effectiveness of our approach, we collect a dataset for 3D reasoning segmentation, comprising over one thousand point-instruction pairs. These pairs are annotated from Matterport3D~\cite{Matterport3D} and ScanNetv2~\cite{dai2017scannet} with implicit text queries that demand complex reasoning knowledge.
%
%

\noindent{The main contributions of this work are:}

\begin{itemize}[align=right,itemindent=0em,labelsep=2pt,labelwidth=1em,leftmargin=*,itemsep=0em] 
\item We introduce Reason3D, a comprehensive framework for reasoning and searching within 3D scenes using extensive language prompts. Reason3D processes 3D point clouds and language inputs to generate both textual outputs and detailed 3D segmentation masks. It supports a wide range of tasks, including expressive 3D referring segmentation, 3D reasoning segmentation, hierarchical 3D searching, and 3D question answering.
\item We establish the novel task of 3D reasoning segmentation, which involves interpreting implicit human instructions within 3D scenes, and we have built a dataset to evaluate this task.
\item We develop a hierarchical mask decoder to effectively address the challenges posed by the sparsity and extensive scale of 3D point clouds.
This approach first identifies a coarse region likely containing the object and uses this region's probability as a prior to guide the refinement of the final mask prediction.
\end{itemize}
\section{Related Work}

\noindent{\bf 3D Point Cloud Segmentation.} 
Recent advancements in point cloud segmentation~\cite{kolodiazhnyi2023oneformer3d,spformer,wu2022point,Schult23ICRA,Wang2023OctFormer} have led to improved class-aware prediction techniques, predominantly employing UNet-like models that process data as either 3D points or voxels. 
Point-based methods~\cite{Chen_HAIS_2021_ICCV,zhao2021point} enhance features with aggregation mechanisms or transformers, while voxel-based methods~\cite{hou2019sis,choy20194d} transform irregular point clouds into regular voxel grids for processing with 3D convolutional networks. 
Another line of work~\cite{Peng2023OpenScene,takmaz2023openmask3d,conceptgraphs,nguyen2023open3dis} focuses on understanding 3D scenes with open-vocabulary inquiry.
As 3D segmentation tasks mature, developing more advanced interactions with these systems using complex instruction has become essential.

The 3D express referring segmentation task~\cite{chen2020scanrefer,achlioptas2020referit_3d} enhances interaction through human language by segmenting 3D objects based on specific textual descriptions. 
TGNN~\cite{huang2021tgnn} uses a two-stage approach to integrate instance and textual features, computing a matching score to identify targets. X-RefSeg3D~\cite{xrefseg3d} combines linguistic and visual features to create a cross-modal scene graph for interactions based on textual and spatial relations. Similarly, 3D-STMN~\cite{wu2024stmn} aligns superpoints with textual inputs to enhance multimodal representation.
However, while these studies make significant strides in object identification using spatial relation cues, they do not fully explore deeper reasoning capabilities. In this work, we introduce Reason3D, a novel approach that extends beyond traditional identification to incorporate advanced reasoning with 3D segmentation models, addressing complex interactions not yet tackled by existing methodologies.
 
\smallskip{\noindent{\bf Large Language Model.}} 
Recent advancements in large language models (LLMs)~\cite{Achiam2023GPT4TR, zheng2023judging, llama, llama2, opt-iml} have showcased their broad generalization across diverse language tasks, thanks to training on extensive textual datasets. 
Through self-supervised learning techniques such as token prediction and masked token reconstruction, as well as further refinements via instruction tuning and specialized datasets, researchers have significantly enhanced the adaptability of these models to new tasks.
Building on this, the remarkable reasoning capabilities of LLMs are increasingly applied in multimodal contexts. Modern models incorporate advanced architectures that integrate visual data~\cite{li2023blip2,Alayrac2022FlamingoAV,awadalla2023openflamingo,liu2023llava,zhu2023minigpt, lai2023lisa, dai2023instructblip,wu2023see,ye2023mplugowl,zhang2023gpt4roi,vstar,qi2024generalizable}, utilizing mechanisms such as cross-attention and image-text feature alignment to enable comprehensive multimodal understanding. 
This has paved the way for models that engage in visual question answering and perform complex reasoning tasks. 
Notably, LISA~\cite{lai2023lisa} introduces a specialized segmentation token into its vocabulary, decoded to generate a segmentation mask, enabling more precise reasoning capabilities. 

Recent efforts have extended large language models (LLMs) to include 3D data for understanding point clouds. 
Point-LLM~\cite{xu2023pointllm} interprets object-level points using LLMs, while 3D-LLM~\cite{3dllm} enhances understanding by integrating multi-view image features with LLMs. LL3DA~\cite{chen2023ll3da} combines textual instructions with visual interactions to improve feature extraction for more effective instruction-following. 
Unlike existing methods, which are limited to bounding box-level grounding, textual responses, or lack contextual reasoning, our approach enables fine-grained segmentation of precisely searched 
objects within 3D data.

\section{3D Reasoning Segmentation}

\begin{figure}[t]
    \centering
    \includegraphics[width=1\linewidth]{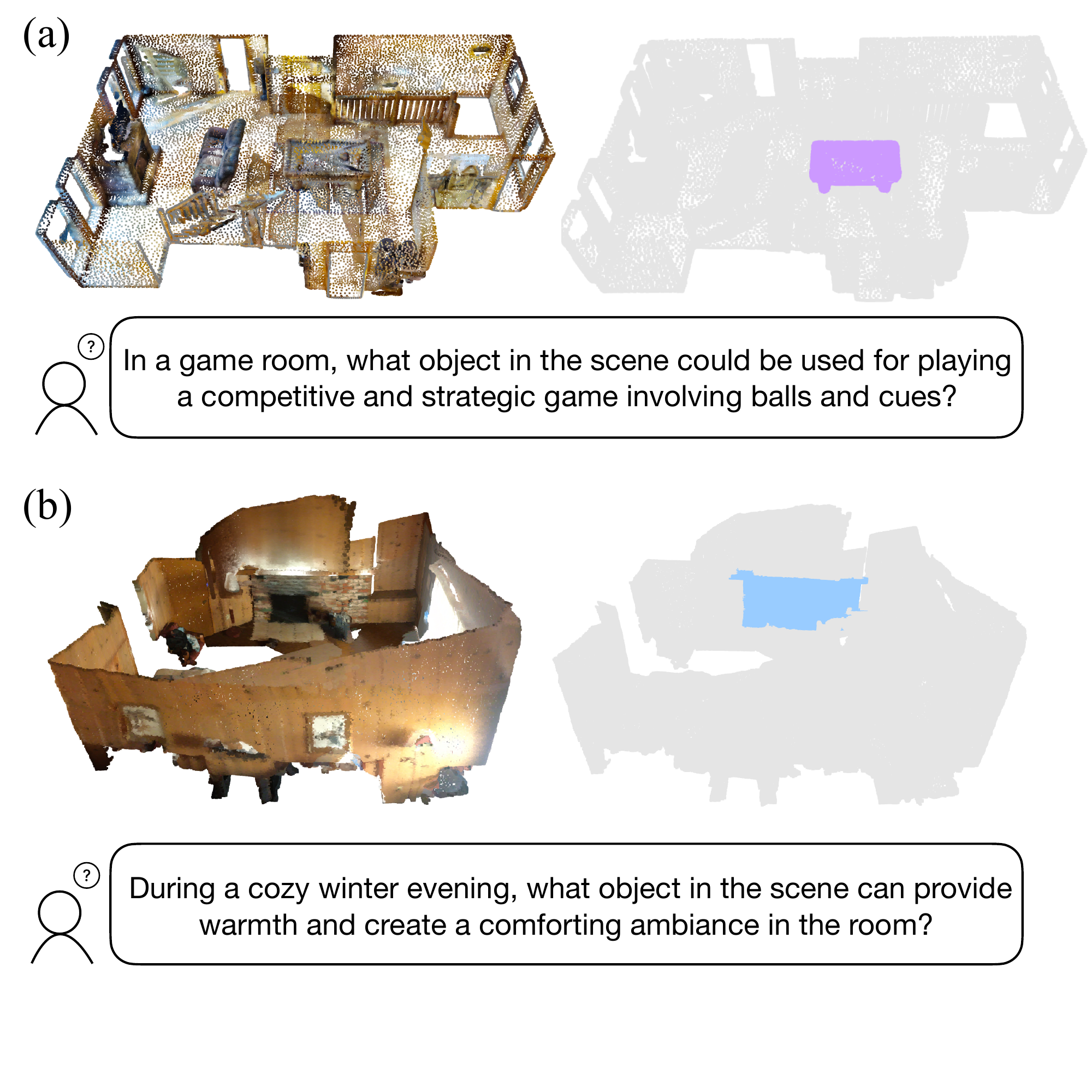}
    \vspace{-1.1cm}
\caption{\textbf{Annotated Sample Examples.} (a) shows a sample from the Matterport3D dataset with the answer \textcolor{plotsamplepurple}{pool table}. (b) presents a sample from the ScannetV2 dataset with the answer \textcolor{plotsampleblue}{fireplace}.}
\label{fig:sample}
\end{figure}

\label{sec:define}
\noindent{\bf Problem Definition.}
3D reasoning segmentation task involves generating a 3D segmentation map $\mathbf{M}$ from a given 3D scene point cloud $\mathbf{P}$ alongside a complex textual instruction $\mathbf{X}_{\rm{txt}}$. This instruction often demands sophisticated linguistic comprehension, extending beyond mere identification tasks, like 3D referring segmentation task~\cite{huang2021tgnn}.
For instance, rather than processing simple directives like "the red chair," the textual queries might involve intricate descriptions or scenarios, such as "an object usually situated in a living room that can accommodate multiple people sitting together comfortably." which requires in-depth world knowledge and reasoning understanding.

\smallskip{\noindent{\bf Dataset Collection.}}
Given the absence of a standardized dataset for evaluating 3D reasoning segmentation, we have collected the 3D scans from indoor datasets, Matterport3D~\cite{Matterport3D} and ScanNetv2~\cite{dai2017scannet} and annotated them with complex text instructions and detailed 3D segmentation masks.  
The dataset consists of 1339 samples for training and 1145 samples for validation.
Two sample data are shown in Figure~\ref{fig:sample}.
More details can be found in the supplementary materials.

\smallskip{\noindent{\bf 3D Hierarchical Searching.}} 
Building on the foundation of 3D reasoning segmentation, we can extend the task to include searching for an object within a specified location in a large-scale 3D scene based on an abstract query. For example, instead of merely finding an object to sit on, we can specify that the object should be inside a bedroom, thus precisely limiting the object's location.

\begin{figure*}[t]
    \centering
    \includegraphics[width=0.975\linewidth]{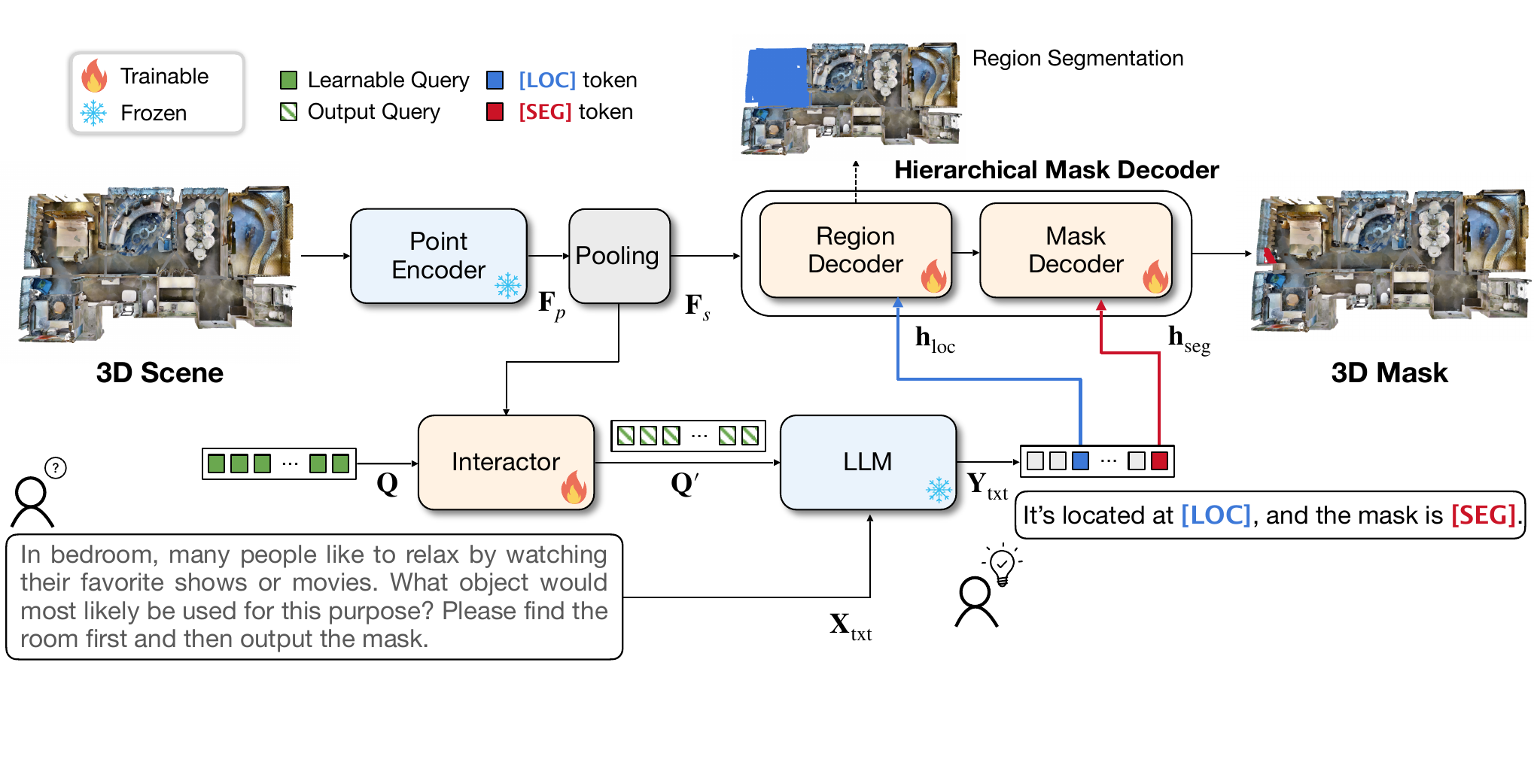}
    \vspace{-1.63cm}
    \caption{\textbf{Overview of our Reason3D framework.} 
    Initially, we utilize a point encoder to extract point features from the input scene, which are simplified by a superpoint pooling layer to reduce complexity. 
    An interactor merges these superpoint features with a learnable query, input into a frozen LLM along with instructions to generate an output containing specifical tokens, \textcolor{plotblue}{{\texttt{[LOC]}}} and \textcolor{plotred}{{\texttt{[SEG]}}}. A hierarchical mask decoder then utilizes the \textcolor{plotblue}{{\texttt{[LOC]}}} embedding to estimate a coarse location that likely covers the target object. Finally, this estimated location prior is integrated with the \textcolor{plotred}{{\texttt{[SEG]}}} embedding to enable the prediction of the final segmentation masks.
    }
\label{fig:arch}
\end{figure*}

\section{Reason3D}
\label{sec:method}


We introduce Reason3D, a novel LLM-based framework for searching and reasoning within 3D point clouds, as illustrated in Figure~\ref{fig:arch}. 
Given a 3D point cloud and a textual query describing an object of interest, our method leverages an LLM model to align point features and predict dense object segmentation masks.
Section~\ref{sub_sec:alignment} discusses the alignment of point clouds with LLMs in the feature space.
Section~\ref{sub_sec:decoder} introduces the proposed hierarchical mask decoder, which employs a coarse-to-fine approach for generating dense segmentation masks.
Finally, Section~\ref{sub_sec:training} details the training loss of our Reason3D framework. 

\subsection{Alignment between LLMs and Point Cloud}
\label{sub_sec:alignment} 
Given a point cloud $\mathbf{P} \in \mathbb{R}^{N \times 6} $ consisting of $N$ points, each characterized by three colors channels (r, g, b) and three coordinates (x, y, z), we aim to extract point features and align them with decoder-only LLM models to facilitate 3D scene understanding based on textual instructions.

\smallskip{\noindent{\bf Scene Encoder.}}
We employ a voxelization operation on the point cloud and utilize a U-Net style backbone~\cite{3dsparse} to extract point-wise features $\mathbf{F}_{p} \in \mathbb{R}^{N \times C}$, where $C$ denotes the channel dimension. 
To further reduce complexity, we feed these features into a superpoint pooling layer that leverages pre-computed superpoints~\cite{superpoint}.
This layer aggregates superpoint features $\mathbf{F}_{s} \in \mathbb{R}^{M \times C}$ by performing average pooling on the point-wise features within each superpoint, effectively reducing the number of points from $N$ to $M$, where $M$ represents the number of superpoints.

This reduction is crucial for managing large-scale scenes without needing to divide the point cloud into smaller segments. For example, a single Matterport3D scene~\cite{Matterport3D} contains approximately one million points, posing a significant challenge for existing algorithms~\cite{Peng2023OpenScene}, which typically require data segmentation. Our approach addresses this challenge by utilizing superpoints, enabling us to handle extensive data in a single pass.

\smallskip{\noindent{\bf Alignment with LLM.}}
To align the superpoint features $\mathbf{F}_{s}$ with existing decoder-only LLM models, we employ an Interactor $\mathcal{F}$ following Q-Former~\cite{li2023blip2} to facilitate dynamic interaction between the point cloud features $\mathbf{F}_{s}$ and the learnable query $\mathbf{Q}$, resulting in an output query $\mathbf{Q}' = \mathcal{F}(\mathbf{Q}, \mathbf{F}_{s})$.
Subsequently, the output query $\mathbf{Q}'$ and textual instructions $\mathbf{X}_{\rm txt}$ are fed into a frozen decoder-only language model (LLM) to generate targeted responses:
\begin{equation}
\rm{\mathbf{Y}_{txt}}=\rm{LLM}(\mathbf{Q}', \mathbf{X}_{txt}).\\
\end{equation}
We freeze the point cloud encoders and the LLM, allowing updates only to the interactor module. 
This setup focuses on learning interactions between 3D and linguistic data, enhancing the model's ability to produce accurate, contextually relevant responses to textual commands about 3D data.

\subsection{Hierarchical Mask Decoder}
\label{sub_sec:decoder} 

Current LLM-based methods for 3D scene understanding \cite{3dllm, chen2023ll3da} are limited to producing textual or numerical outputs and cannot thus predict dense 3D masks. To overcome these limitations, we propose a Hierarchical Mask Decoder (HMD) that utilizes a coarse-to-fine approach to predict segmentation tasks guided by the output of the LLM.
Specifically, we first utilize a location prompt $\mathbf{P}_{\rm{loc}}$ to learn the coarse location of the mask that potentially covers the target object. This coarse location then serves as a prior for learning the object segmentation mask, guided by another segmentation prompt $\mathbf{P}_{\rm{seg}}$. We will provide a more detailed explanation later.

The Hierarchical Mask Decoder predicts the segmentation masks $\mathbf{M}_{\rm {seg}}$ by utilizing the superpoint features $\mathbf{F}_s$ and two specific prompts $\langle \mathbf{P}_{\rm {loc}}, \mathbf{P}_{\rm {seg}} \rangle$, which is based on the instruction $\mathbf{X}_{\rm txt}$:
\begin{align}
    \mathbf{M}_{\rm {seg}} = 
    {\rm HMDecoder} (\mathbf{F}_s; \langle \mathbf{P}_{\rm {loc}}, \mathbf{P}_{\rm {seg}} \rangle | \langle \mathbf{F}_s, \mathbf{X}_{\rm {txt}} \rangle).
    \label{eq:prompt}
\end{align} 
To generate these two prompt features for guidance, we direct the LLM to output specific embedding tokens inspired by LISA \cite{lai2023lisa}.
Considering the segmentation prompt, the LLM is instructed to generate a \textcolor{plotred}{{\texttt{[SEG]}}} token. The last-layer embedding $\mathbf{h}_{\rm{seg}}$ associated with the \textcolor{plotred}{{\texttt{[SEG]}}} token is transformed through an MLP projection layer $\mathcal{G}$, resulting in the segmentation prompt $\mathbf{P}_{\rm{seg}}=\mathcal{G}({\mathbf{h}}_{\rm{seg}})$. This prompt encapsulates the features of the target object, derived from the textual instructions, to guide the final mask prediction.

Moreover, to effectively target small objects within large scenes, we introduce a location token, \textcolor{plotblue}{{\texttt{[LOC]}}}, which the LLM is instructed to generate. This token learns a coarse location that may potentially encompass the object mask, serving as a prior feature to enhance the accuracy of the final segmentation results.
Similar to the \textcolor{plotred}{{\texttt{[SEG]}}} token process, we refine the embedding $\mathbf{h}_{\rm{loc}}$ of the \textcolor{plotblue}{{\texttt{[LOC]}}} token using an MLP layer $\mathcal{G}$, resulting in the location prompts $\mathbf{P}_{\rm{loc}}=\mathcal{G}({\mathbf{h}}_{\rm{loc}})$.
To this end, we can use these two prompts to guide the final mask prediction. 
In practice, we first exploit a region decoder, $\mathcal{F}_{\rm{loc}}$, built on a transformer decoder architecture~\cite{Schult23ICRA, spformer}. The location prompt $\mathbf{P}_{\rm{loc}}$ serves as the query, while the superpoint features $\mathbf{F_{s}}$ act as key and value to generate the location mask $\mathbf{M}_{\rm{loc}}$:
\begin{align}
\mathbf{M}_{\rm{loc}} = \mathcal{F}_{\rm{loc}}({\mathbf{P}}_{\rm{loc}}, \mathbf{F_{s}}),
\end{align}
where the location mask indicates the probability of the region potentially covering the target object.

After that, we use an MLP layer $\mathcal{H}_{\rm{loc}}$ to encode the location mask $\mathbf{M}_{\rm{loc}}$, serving as a feature prior, which is then combined with the point features.
Subsequently, we make the final mask generation based on the segmentation prompt and the integrated point features:
\begin{align}
\mathbf{M}_{\rm{seg}} = \mathcal{F}_{\rm{seg}}({\mathbf{P}}_{\rm{seg}}, \mathbf{F_{s}}+\mathcal{H}_{\rm{loc}}(\mathbf{M}_{\rm{loc}})),
\label{eq:m_seg}
\end{align}
where $\mathbf{M}_{\rm{seg}}$ is the final mask, and $\mathcal{F}_{\rm{seg}}$ shares the same architectural framework as $\mathcal{F}_{\rm{loc}}$.

\subsection{Training Reason3D} 
\label{sub_sec:training} 
The loss function for Reason3D comprises two essential components: the LLM loss $\mathcal{L}_{\rm{llm}}$ and the segmentation mask loss $\mathcal{L}_{\rm{mask}}$. The overall combination is represented as:
\begin{equation}
    \mathcal{L} = \mathcal{L}_{\rm{llm}} +\mathcal{L}_{\rm{mask}}.
\end{equation}
In particular, The LLM loss, $\mathcal{L}_{\rm{llm}}$, embodies the linguistic aspects through an auto-regressive cross-entropy loss for text generation, incorporating cross-entropy loss ${\rm CE}$ for each token: 
\begin{align}
\begin{aligned}
    \mathcal{L}_{\rm{llm}} & = {\rm CE}(\mathbf{Y}_{\rm{txt}}, \hat{\mathbf{Y}}_{\rm{txt}})
\end{aligned}
\end{align}
where $\hat{\mathbf{Y}}_{\rm{txt}}$ represents the ground truth word token. 
In addition, the mask loss $\mathcal{L}_{\rm{mask}}$ aims at encouraging the model to generate high-quality segmentation masks. This loss is computed using a binary cross-entropy (BCE) loss and DICE loss for all superpoints, which is represented as:
\begin{align}
\begin{aligned}
    \mathcal{L}_{\rm{mask}_{*}} = {\rm BCE}&(\mathbf{M}_{*}, \hat{\mathbf{M}_{*}})  + {\rm DICE}(\mathbf{M}_{*}, \hat{\mathbf{M}}_{*}), * \in [\rm{loc}, \rm{seg}].
\end{aligned}
\end{align}
where $\hat{\mathbf{M}}_{*}$ means the ground truth segmentation mask for region-level and object-level superpoints. 
For the object-level mask $\hat{\mathbf{M}}_{\rm{seg}}$, we use the mask corresponding to the specific object we are targeting. For the region-level mask $\hat{\mathbf{M}}_{\rm{loc}}$, we designate the points as foreground points if the distance between any point and the object's center is smaller than threshold $\tau$ or we select the points within the specific room for hierarchical searching task.

\section{Experiments}

\begin{table*}[t]
\footnotesize
\centering
\setlength\tabcolsep{8pt}
\begin{tabular}{lccccccc}
\toprule
\multirow{2}*{\textbf{Method}} & \multirow{2}*{\textbf{Venue}} & \multicolumn{3}{c}
{\textbf{ScanNet}} & \multicolumn{3}{c}{\textbf{Matterport3D}}  \\
\cmidrule(lr){3-5} \cmidrule(lr){6-8} 
&& Acc@0.25 & Acc@0.50 & mIoU & Acc@0.25 & Acc@0.50 & mIoU \\
\midrule
OpenScene~\cite{Peng2023OpenScene} & CVPR'23 & 4.22 & 0.97 & 5.03 & 4.07 & 0.57 & 6.36\\ 
OpenScene~\cite{Peng2023OpenScene}+FlanT5~\cite{flant5} & CVPR'23+ArXiv'22 & 24.68 & 7.14 & 15.03 & 19.98 & 4.02& 13.60\\
OpenMask3D~\cite{takmaz2023openmask3d} & NeurIPS'23 & 5.70 & 3.25 & 7.14 & 3.25 & 0.12 & 5.96\\
OpenMask3D~\cite{takmaz2023openmask3d}+FlanT5~\cite{flant5}& NeurIPS'23+ArXiv'22 & 20.78 & 6.82 & 13.38 & 17.46 & 0.23 & 9.07 \\ \midrule
3D-STMN~\cite{wu2024stmn} & AAAI'24 & 25.43 & 17.78 & 18.23 & 20.68 & 10.81 & 13.47 \\ 
Llama2~\cite{llama2}+CLIP~\cite{clip} & ArXiv'23+ArXiv'22 & 39.26 & 25.93 & 27.23 & 28.51 & 14.86 & 17.80\\ 
\midrule
Reason3D (Ours) & - & \textbf{43.21} & \textbf{32.10} &
\textbf{31.20} & \textbf{31.22} & \textbf{17.43} & \textbf{19.54} \\
\bottomrule
\end{tabular}
\vspace{-2mm}
\caption{\textbf{3D Reasoning Segmentation Results.}
The evaluation metric is accuracy at IoU 0.25, IoU 0.5 and mIoU.}
\label{tab:results_reason}
\end{table*}

\begin{table*}[t]
\footnotesize
\centering
\setlength\tabcolsep{3.7pt}
\begin{tabular}{llccccccccc}
\toprule &
\multirow{2}*{\textbf{Method}}  & \multicolumn{3}{c}
{\textbf{Room Num = 1$\sim$2}} & \multicolumn{3}{c}{\textbf{Room Num = 3$\sim$4}} & \multicolumn{3}{c}{\textbf{Room Num $\ge$ 5}} \\
\cmidrule(lr){3-5} \cmidrule(lr){6-8} \cmidrule(lr){9-11}
&&Acc@0.25 & Acc@0.50 & mIoU & Acc@0.25 & Acc@0.50 & mIoU & Acc@0.25 & Acc@0.50 & mIoU \\
\midrule
(a) & FlanT5~\cite{flant5} + OpenScene~\cite{Peng2023OpenScene} & 17.65 & 3.95 & 12.89 & 11.27 & 1.02 & 7.69 & 6.22 & 0.97 & 2.21 \\
\midrule
(b) & Reason3D-base & 25.23 & 10.32 & 15.56 & 12.84 & 5.50 & 8.23 & 8.26 & 2.52 & 5.33\\ 
(c) & Region Seg + Reason3D-base & 26.98 & 13.21 & 17.02 & 19.21 & 8.21 & 11.67 & 11.96 & 4.78 & 7.21 \\
(d) & Reason3D & \textbf{29.82} & \textbf{16.97} & \textbf{18.81} & \textbf{22.25} & \textbf{11.93} & \textbf{14.12} & \textbf{16.06} & \textbf{7.34} & \textbf{10.35} \\
\bottomrule
\end{tabular}
\vspace{-2mm}
\caption{\textbf{3D Hierarchical Searching Results} on Matterport3D dataset with different room numbers.
Reason3D-base refers to the full Reason3D model without the proposed [LOC] token and region decoder. The evaluation metric is IoU@0.25, IoU@0.5 and mIoU.}
\label{tab:results_search}
\end{table*}

\begin{table*}[t]
\footnotesize
\centering
\setlength\tabcolsep{5pt}
\begin{tabular}{lcccccccccc}
\toprule
\multirow{2}*{\textbf{Method}} & \multirow{2}*{\textbf{Venue}} & \multicolumn{3}{c}
{\textbf{Unique ($\sim$19\%)}} & \multicolumn{3}{c}{\textbf{Multiple ($\sim$81\%)}} & \multicolumn{3}{c}{\textbf{Overall}} \\
\cmidrule(lr){3-5} \cmidrule(lr){6-8} \cmidrule(lr){9-11}
&& Acc@0.25 & Acc@0.50 & mIoU & Acc@0.25 & Acc@0.50 & mIoU & Acc@0.25 & Acc@0.50 & mIoU \\
\midrule
ScanRefer~\cite{chen2020scanrefer}* & ECCV'20 & 67.6 & 44.4 & 39.9 & 31.2 & 20.9 & 19.5 & 38.2 & 25.5 & 23.5 \\ 
3DVG-Transformer~\cite{zhao2021_3DVG_Transformer}* & ICCV'21 & 79.5 & 58.0 & 49.9 & 42.0 & 30.8 & 27.0 & 49.3 & 36.1 & 31.4 \\
3D-SPS~\cite{luo20223d}* & CVPR'22 & 84.8 & 65.6 & 54.7 & 41.7 & 30.8 & 26.7 & 50.1 & 37.6 & 32.1 \\
3D-LLM~\cite{3dllm}* & NeurIPS'23 & 57.8 & 30.6 & 32.5 & 24.7 & 12.8 & 14.0 & 31.1 & 16.3 & 17.6 \\
\midrule
TGNN~\cite{huang2021tgnn} & AAAI'21 & 69.3 & 57.8 & 50.7 & 31.2 & 26.6 & 23.6 & 38.6 & 32.7 & 28.8 \\
X-RefSeg3D~\cite{xrefseg3d} & AAAI'24 & - & - & - & - & - & - & 40.3 & 33.8 & 29.9\\
3D-STMN~\cite{wu2024stmn} & AAAI'24 & \textbf{89.3} & 84.0 & 74.5 & 46.2 & 29.2 & 31.1 & 54.6 & 39.8 & 39.5  \\
\midrule
Reason3D (Ours) & - & 88.4 & \textbf{84.2} & \textbf{74.6} & \textbf{50.5} & \textbf{31.7} & \textbf{34.1} & \textbf{57.9} & \textbf{41.9} & \textbf{42.0} \\
\bottomrule
\end{tabular}
\vspace{-2mm}
\caption{\textbf{3D Referring Expression Segmentation Results} on ScanRefer dataset with the accuracy evaluated by IoU 0.25, IoU 0.5 and mIoU. 
For the first block methods * that only output 3D bounding boxes, we reproduce the results based on their official codes by extracting the points inside the boxes as the segmentation mask predictions.}
\label{tab:results_refer}
\end{table*}

\setlength{\tabcolsep}{0.03\linewidth}{
\begin{table}[t]
    \footnotesize
    \centering
    \begin{tabular}{l  @{\hspace{4pt}} l @{\hspace{6pt}} c @{\hspace{6pt}} c}
        \toprule
        & Ablation & Acc@0.25 & Acc@0.50 \\ 
        \midrule
        (a) & w/o region sup. ($\mathcal{L}_{\rm{mask}_{loc}}$) & 14.27 & 6.33  \\ %
        (b) & hard thresholding for $\mathbf{M}_{\rm{loc}}$ & 20.35 & 9.98   \\  
        (c) & Eq.~\ref{eq:m_seg} $\rightarrow$ concat. $\mathcal{F}_{\rm{seg}}$ and $\mathcal{H}_{\rm{loc}}(\mathbf{M}_{\rm{loc}})$& 21.53 & 10.98\\
        (d) & Reason3D (full model) & 22.25 & 11.93 \\ %
        \bottomrule
    \end{tabular}
    \vspace{-2mm}
    \caption{\textbf{The effect of different designs in Hierarchical Mask Decoder (HMD)} on Matterport dataset. 
    We use room number = 3$\sim$4 as the main experiments.
    }
    \label{tab:abl_hmd}
\end{table}
}

\subsection{Experimental Setting}

\noindent{\bf Datasets.} 
Our training data includes three main types of datasets:
(1) For the 3D expressive referring segmentation task, we use ScanRefer~\cite{chen2020scanrefer} and Sr3D datasets~\cite{achlioptas2020referit_3d}.
(2) For the 3D question answering task, we utilize ScanQA dataset~\cite{azuma_2022_CVPR}.
(3) For the 3D reasoning segmentation task, we construct the Reason3D dataset from ScanNetV2 and Matterport3D datasets. The results of 3D QA and more details are included in the supplementary materials.

\smallskip{\noindent{\bf Model Architecture.}} 
We use a pre-trained Sparse 3D U-Net~\cite{spformer} to extract point-wise features. For the language learning model, we employ FlanT5~\cite{flant5}, maintaining most of its pre-trained weights frozen, except for adapting the weights for the newly-added location and segmentation tokens. Our Interactor is constructed following BLIP-2~\cite{li2023blip2}, incorporating 1408-dimensional features.

\smallskip{\noindent{\bf Evaluation Metrics.}} 
For the 3D expressive referring segmentation and 3D reasoning segmentation tasks, the primary evaluation metrics are Mean Intersection over Union (mIoU), which quantifies the average overlap between the predicted and true 3D volumes, and Accuracy at k Intersection over Union (Acc@kIoU). This latter metric measures the proportion of descriptions for which the predicted mask overlaps the ground truth with an IoU greater than k, where k is set at thresholds of 0.25 and 0.5, thus assessing the model's performance at varying levels of precision.
%


\subsection{3D Reasoning Segmentation Results}
Table~\ref{tab:results_reason} presents the results of 3D reasoning segmentation, where our model significantly outperforms previous methods, achieving a notable increase in mean Intersection over Union (mIoU). Unlike typical 3D referring segmentation tasks, this task demands not only spatial understanding but also robust reasoning and contextual comprehension.

Our model excels at interpreting long sentence queries and managing 3D reasoning segmentation tasks, outperforming open-vocabulary segmentation methods like OpenScene~\cite{Peng2023OpenScene} and OpenMask3D~\cite{takmaz2023openmask3d}, which primarily use vocabulary as the query.
We also compare it to the two-stage methods, where FlanT5~\cite{flant5} generates a short vocabulary output followed by segmentation with OpenScene~\cite{Peng2023OpenScene} or OpenMask3D~\cite{takmaz2023openmask3d}.  
Our approach surpasses these by leveraging more expressive hidden embeddings, offering a richer representation than relying solely on text as an intermediary.

Compared to leading 3D referring segmentation models like 3D-STMN~\cite{wu2024stmn} fine-tuned on the Reason3D dataset, we find that while 3D-STMN excels in direct referencing, it struggles with indirect queries. In contrast, our model, with its integration of large language models, shows superior adaptability and performance in these scenarios.

We also compare our model to a two-stage method that combines Llama2~\cite{llama2} and CLIP~\cite{clip}, both fine-tuned on the Reason3D dataset. In this approach, Llama2~\cite{llama2} generates a vocabulary output, which CLIP~\cite{clip} converts into textual features. These features then interact with the same point features and mask decoder used in our Reason3D to produce segmentation masks.
The results show that our approach significantly outperforms this two-stage method, which is fully decoupled and relies solely on the textual outputs from the LLM.

\subsection{3D Hierarchical Searching Results}
For the 3D hierarchical searching task, an extension of 3D reasoning segmentation task, our goal is to segment target objects within a larger space (\eg, multiple rooms) rather than a single room.
The task involves specifying a target room where the model must locate the object, such as finding the TV in the bedroom, as shown in Figure~\ref{fig:arch}.
Given that the Reason3D dataset mainly focuses on single-room scenarios, we extend it to multi-room settings by reusing a subset of annotated Matterport3D~\cite{Matterport3D} data. We chose not to use ScanNet, as it features only single-room scenes.

Table~\ref{tab:results_search} presents different results: (a) a two-stage baseline that uses an LLM~\cite{flant5} to parse the instruction, and applies an open-world segmentation model~\cite{Peng2023OpenScene}; (b) Reason3D-base (without Region Decoder), (c) Reason3D combined with a region segmentation model that first segments the target room's region, then applies the Reason3D-base for the segmented region (without location prompt), and (d) Reason3D (full model).
Reason3D (d) outperforms the two-stage baseline (a) due to its effective design.
Comparing (b) and (d) shows that the Hierarchical Mask Decoder (HMD) significantly boosts performance by using an additional token to guide coarse region learning, effectively managing segmentation masks, especially as point cloud complexity increases.
Additionally, baseline (c) improves upon (b) but faces optimization challenges in the two-stage training pipeline compared to the full model (d).

\subsection{3D Referring Expression Segmentation Results}
To demonstrate the effectiveness of our model in the 3D express referring segmentation task, we compare Reason3D against state-of-the-art methods on the ScanRefer validation set, as shown in Table~\ref{tab:results_refer}. 
Our approach significantly outperforms 3D-STMN~\cite{wu2024stmn} in overall performance. Given the limited focus on 3D referring segmentation in the literature, we also compare our model to several 3D grounding approaches that predict only 3D bounding boxes, which can generate segmentation masks by extracting points within the predicted boxes.  
Notably, our approach vastly outperforms the LLM-based method, 3D-LLM~\cite{3dllm}, which struggles with accurately locating 3D boxes and effectively extracting segmentation masks.

\subsection{Ablation Study}
\smallskip{
\noindent{\bf Effectiveness of different design in HMD.}}
Besides the Reason3D-base (without the proposed coarse-to-fine approach) in Table~\ref{tab:results_search}, we also conduct an ablation study on the Matterport3D dataset for the 3D hierarchical searching task to validate the impact of various designs in our Hierarchical Mask Decoder (HMD), as shown in Table \ref{tab:abl_hmd}. 
When the region supervision term ($\mathcal{L}_{\rm{mask}_{loc}}$) is removed (a), the model's performance drops significantly, achieving only 14.27\% accuracy at a 0.25 IoU, which indicates that region supervision plays a crucial role in guiding the HMD decoder to learn the priors for segmentation predictions. 
Comparing (b) and (d), we observe that hard thresholding performs worse than probability-based region mask prediction. Hard thresholding discards uncertainty, while using probability as a prior retains valuable information, enabling more nuanced and accurate segmentation decisions. 
Furthermore, comparing (c) and (d) suggests that summation is more effective than concatenation for combining point features with learned location priors in Equation~\ref{eq:m_seg}, likely due to its more integrated and simpler feature representation.

\begin{table*}[t]
\vspace{-.2em}
\centering
\hspace{-0.5em}
\subfloat[
{Superpoint Pooling}.
\label{tab:decoder_depth}
]{
\centering
\begin{minipage}{0.3\linewidth}{
\begin{center}
\tablestyle{2pt}{1.00}
\scriptsize
\begin{tabular}{cccc} \toprule
Superpoint & Pool & Acc@0.25 & Time (ms) \\
\midrule
\xmark & - & 37.55 & 486.1 \\
\cmark & max & 42.97 & 271.3 \\
\cmark & Avg & 43.21 & 268.5 \\
\bottomrule
\end{tabular}
\end{center}}\end{minipage}
}
\hspace{1.2em}
\subfloat[
Segmentation Loss.
\label{tab:decoder_width}
]{
\begin{minipage}{0.27\linewidth}{\begin{center}
\tablestyle{2pt}{1.00}
\scriptsize
\begin{tabular}{cccc}
\toprule
{\rm DICE} & {\rm BCE} & Acc@0.25 & Acc@0.50 \\
\midrule
\cmark &  & 41.73  & 31.36 \\
 & \cmark & 30.86 & 22.47  \\
\cmark & \cmark & 43.21 & 32.10  \\
\bottomrule
\end{tabular}
\end{center}}\end{minipage}
}
\hspace{0.1em}
\subfloat[
Layer of Decoders.
\label{tab:mask_token}
]{
\begin{minipage}{0.29\linewidth}{\begin{center}
\tablestyle{2pt}{1.0}
\scriptsize
\begin{tabular}{ccc}
\toprule
Layer & Acc@0.25 & Acc@0.50 \\
\midrule
1 & 40.25 & 27.65  \\ 
3 & 42.78 & 30.62  \\ 
6 & 43.21 & 32.10   \\
\bottomrule
\end{tabular}
\end{center}}\end{minipage}}
\\
\centering
\vspace{-0.7em}
\caption{\textbf{Ablation experiments} for different design on the scannetV2 dataset for the 3D reasoning segmentation task.} \vspace{-0.5em}
\label{tab:ablations}
\end{table*}

\smallskip{
\noindent{\bf Effectiveness of superpoints.}}
Table~\ref{tab:ablations}(a) proves that the superpoints pooling operation is essential for our pipeline since it helps to reduce the training complexity and enables the effective training of the pipeline. Also, the average pooling for the superpoints can achieve better performance.

\smallskip{
\noindent{\bf Effectiveness of different segmentation loss.}}
Table~\ref{tab:ablations}(b) presents the performance impact of various components of segmentation loss. Using either Binary Cross-Entropy (BCE) loss or Dice loss alone leads to significantly reduced performance. In contrast, combining Dice loss and BCE loss results in the most favorable outcomes.

\smallskip{
\noindent{\bf Effectiveness of different decoder layers.}}
Table~\ref{tab:ablations}(c) presents the impact of different numbers of decoder layers. We use six layers for the decoder as the default number.

\subsection{Visualization Results}
\begin{figure*}[t]
    \centering
    \includegraphics[width=0.89\linewidth]{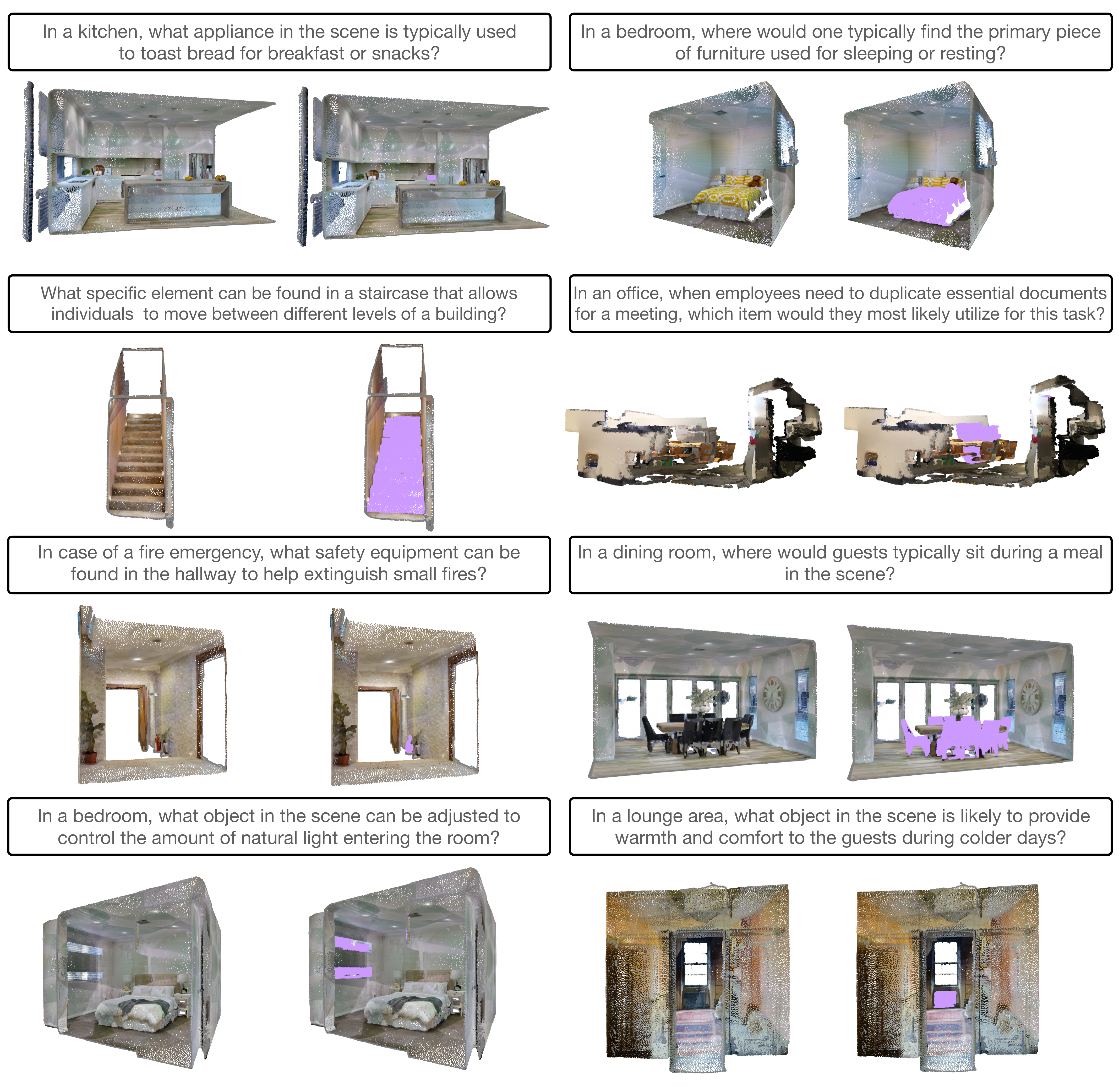}
    \vspace{-0.4cm}    \caption{\textbf{Visualization Results for 3D Reasoning Segmentation Tasks.} Each sub-figure presents a textual query alongside the input point cloud. The \textcolor{plotsamplepurple}{purple} regions highlight the predicted segmentation masks generated by our model.}
    \label{fig:vis}
\end{figure*}

Figure~\ref{fig:vis} displays the visualization results of our Reason3D model for the 3D reasoning segmentation task, highlighting our model's proficiency in accurately generating segmentation masks based on the query. Additional visualization results are included in the supplementary materials.

\section{Conclusion}
This paper presents Reason3D, a framework that leverages Large Language Models (LLMs) for enhanced scene understanding, generating textual responses and segmentation predictions. We introduce the novel task of 3D reasoning segmentation, requiring interpreting implicit human instructions within three-dimensional scenes. 
A hierarchical mask decoder is proposed to enhance mask prediction by first identifying a broad region likely to contain the target object, which then serves as feature priors for further refinement. 
Extensive experiments on the ScanNetV2 and Matterport3D datasets demonstrate outstanding performance across tasks like 3D reasoning segmentation, 3D hierarchical searching, 3D referring segmentation, and question answering.

{
    \small
    \bibliographystyle{ieeenat_fullname}
    \bibliography{ref}

\begin{thebibliography}{65}
\providecommand{\natexlab}[1]{#1}
\providecommand{\url}[1]{\texttt{#1}}
\expandafter\ifx\csname urlstyle\endcsname\relax
  \providecommand{\doi}[1]{doi: #1}\else
  \providecommand{\doi}{doi: \begingroup \urlstyle{rm}\Url}\fi

\bibitem[Achlioptas et~al.(2020)Achlioptas, Abdelreheem, Xia, Elhoseiny, and Guibas]{achlioptas2020referit_3d}
Panos Achlioptas, Ahmed Abdelreheem, Fei Xia, Mohamed Elhoseiny, and Leonidas Guibas.
\newblock Referit3d: Neural listeners for fine-grained 3d object identification in real-world scenes.
\newblock In \emph{ECCV}, 2020.

\bibitem[Alayrac et~al.(2022)Alayrac, Donahue, Luc, Miech, Barr, Hasson, Lenc, Mensch, Millican, Reynolds, Ring, Rutherford, Cabi, Han, Gong, Samangooei, Monteiro, Menick, Borgeaud, Brock, Nematzadeh, Sharifzadeh, Binkowski, Barreira, Vinyals, Zisserman, and Simonyan]{Alayrac2022FlamingoAV}
Jean-Baptiste Alayrac, Jeff Donahue, Pauline Luc, Antoine Miech, Iain Barr, Yana Hasson, Karel Lenc, Arthur Mensch, Katie Millican, Malcolm Reynolds, Roman Ring, Eliza Rutherford, Serkan Cabi, Tengda Han, Zhitao Gong, Sina Samangooei, Marianne Monteiro, Jacob Menick, Sebastian Borgeaud, Andy Brock, Aida Nematzadeh, Sahand Sharifzadeh, Mikolaj Binkowski, Ricardo Barreira, Oriol Vinyals, Andrew Zisserman, and Karen Simonyan.
\newblock Flamingo: a visual language model for few-shot learning.
\newblock In \emph{NeurIPS}, 2022.

\bibitem[Awadalla et~al.(2023)Awadalla, Gao, Gardner, Hessel, Hanafy, Zhu, Marathe, Bitton, Gadre, Sagawa, Jitsev, Kornblith, Koh, Ilharco, Wortsman, and Schmidt]{awadalla2023openflamingo}
Anas Awadalla, Irena Gao, Josh Gardner, Jack Hessel, Yusuf Hanafy, Wanrong Zhu, Kalyani Marathe, Yonatan Bitton, Samir Gadre, Shiori Sagawa, Jenia Jitsev, Simon Kornblith, Pang~Wei Koh, Gabriel Ilharco, Mitchell Wortsman, and Ludwig Schmidt.
\newblock Openflamingo: An open-source framework for training large autoregressive vision-language models.
\newblock \emph{arXiv preprint arXiv:2308.01390}, 2023.

\bibitem[Azuma et~al.(2022)Azuma, Miyanishi, Kurita, and Kawanabe]{azuma_2022_CVPR}
Daichi Azuma, Taiki Miyanishi, Shuhei Kurita, and Motoaki Kawanabe.
\newblock Scanqa: 3d question answering for spatial scene understanding.
\newblock In \emph{CVPR}, 2022.

\bibitem[Banerjee and Lavie(2005)]{banarjee2005}
Satanjeev Banerjee and Alon Lavie.
\newblock {METEOR}: An automatic metric for {MT} evaluation with improved correlation with human judgments.
\newblock In \emph{ACL Workshop}, 2005.

\bibitem[Chang et~al.(2017)Chang, Dai, Funkhouser, Halber, Niessner, Savva, Song, Zeng, and Zhang]{Matterport3D}
Angel Chang, Angela Dai, Thomas Funkhouser, Maciej Halber, Matthias Niessner, Manolis Savva, Shuran Song, Andy Zeng, and Yinda Zhang.
\newblock {Matterport3D}: Learning from {RGB-D} data in indoor environments.
\newblock In \emph{3DV}, 2017.

\bibitem[Chen et~al.(2020)Chen, Chang, and Nie{\ss}ner]{chen2020scanrefer}
Dave~Zhenyu Chen, Angel~X Chang, and Matthias Nie{\ss}ner.
\newblock Scanrefer: 3d object localization in rgb-d scans using natural language.
\newblock In \emph{ECCV}, 2020.

\bibitem[Chen et~al.(2021)Chen, Fang, Zhang, Liu, and Wang]{Chen_HAIS_2021_ICCV}
Shaoyu Chen, Jiemin Fang, Qian Zhang, Wenyu Liu, and Xinggang Wang.
\newblock Hierarchical aggregation for 3d instance segmentation.
\newblock In \emph{ICCV}, 2021.

\bibitem[Chen et~al.(2022)Chen, Tapaswi, Guhur, Schmid, and Laptev]{chen2022vil3dref}
Shizhe Chen, Makarand Tapaswi, Pierre-Louis Guhur, Cordelia Schmid, and Ivan Laptev.
\newblock Language conditioned spatial relation reasoning for 3d object grounding.
\newblock In \emph{NeurIPS}, 2022.

\bibitem[Chen et~al.(2024)Chen, Chen, Zhang, Li, Yu, Fei, Zhu, Fan, and Chen]{chen2023ll3da}
Sijin Chen, Xin Chen, Chi Zhang, Mingsheng Li, Gang Yu, Hao Fei, Hongyuan Zhu, Jiayuan Fan, and Tao Chen.
\newblock Ll3da: Visual interactive instruction tuning for omni-3d understanding, reasoning, and planning.
\newblock In \emph{CVPR}, 2024.

\bibitem[Choy et~al.(2019)Choy, Gwak, and Savarese]{choy20194d}
Christopher Choy, JunYoung Gwak, and Silvio Savarese.
\newblock 4d spatio-temporal convnets: Minkowski convolutional neural networks.
\newblock In \emph{CVPR}, 2019.

\bibitem[Chung et~al.(2022)Chung, Hou, Longpre, Zoph, Tay, Fedus, Li, Wang, Dehghani, Brahma, and et~al.]{flant5}
Hyung~Won Chung, Le Hou, Shayne Longpre, Barret Zoph, Yi Tay, William Fedus, Eric Li, Xuezhi Wang, Mostafa Dehghani, Siddhartha Brahma, and et al.
\newblock Scaling instruction-finetuned language models.
\newblock \emph{arXiv preprint arXiv:2210.11416}, 2022.

\bibitem[Dai et~al.(2017)Dai, Chang, Savva, Halber, Funkhouser, and Nie{\ss}ner]{dai2017scannet}
Angela Dai, Angel~X. Chang, Manolis Savva, Maciej Halber, Thomas Funkhouser, and Matthias Nie{\ss}ner.
\newblock Scannet: Richly-annotated 3d reconstructions of indoor scenes.
\newblock In \emph{CVPR}, 2017.

\bibitem[Dai et~al.(2023)Dai, Li, Li, Tiong, Zhao, Wang, Li, Fung, and Hoi]{dai2023instructblip}
Wenliang Dai, Junnan Li, Dongxu Li, Anthony Meng~Huat Tiong, Junqi Zhao, Weisheng Wang, Boyang Li, Pascale Fung, and Steven Hoi.
\newblock Instructblip: Towards general-purpose vision-language models with instruction tuning.
\newblock \emph{arXiv preprint arXiv:2305.06500}, 2023.

\bibitem[Ester et~al.(1996)Ester, Kriegel, Sander, and Xu]{dbscan}
Martin Ester, Hans-Peter Kriegel, Jörg Sander, and Xiaowei Xu.
\newblock A density-based algorithm for discovering clusters in large spatial databases with noise.
\newblock In \emph{KDD}, 1996.

\bibitem[Felzenszwalb and Huttenlocher(2004)]{super}
Pedro~F Felzenszwalb and Daniel~P Huttenlocher.
\newblock Efficient graph-based image segmentation.
\newblock \emph{IJCV}, 2004.

\bibitem[Graham et~al.(2018)Graham, Engelcke, and van~der Maaten]{3dsparse}
Benjamin Graham, Martin Engelcke, and Laurens van~der Maaten.
\newblock 3d semantic segmentation with submanifold sparse convolutional networks.
\newblock In \emph{CVPR}, 2018.

\bibitem[Gu et~al.(2023)Gu, Kuwajerwala, Morin, Jatavallabhula, Sen, Agarwal, Rivera, Paul, Ellis, Chellappa, Gan, {de Melo}, Tenenbaum, Torralba, Shkurti, and Paull]{conceptgraphs}
Qiao Gu, Alihusein Kuwajerwala, Sacha Morin, {Krishna Murthy} Jatavallabhula, Bipasha Sen, Aditya Agarwal, Corban Rivera, William Paul, Kirsty Ellis, Rama Chellappa, Chuang Gan, {Celso Miguel} {de Melo}, {Joshua B.} Tenenbaum, Antonio Torralba, Florian Shkurti, and Liam Paull.
\newblock Conceptgraphs: Open-vocabulary 3d scene graphs for perception and planning.
\newblock \emph{arXiv preprint arXiv:2309.16650}, 2023.

\bibitem[Hong et~al.(2023)Hong, Zhen, Chen, Zheng, Du, Chen, and Gan]{3dllm}
Yining Hong, Haoyu Zhen, Peihao Chen, Shuhong Zheng, Yilun Du, Zhenfang Chen, and Chuang Gan.
\newblock 3d-llm: Injecting the 3d world into large language models.
\newblock In \emph{NeurIPS}, 2023.

\bibitem[Hou et~al.(2019)Hou, Dai, and Nie{\ss}ner]{hou2019sis}
Ji Hou, Angela Dai, and Matthias Nie{\ss}ner.
\newblock 3d-sis: 3d semantic instance segmentation of rgb-d scans.
\newblock In \emph{CVPR}, 2019.

\bibitem[Huang et~al.(2021)Huang, Lee, Chen, and Liu]{huang2021tgnn}
Pin-Hao Huang, Han-Hung Lee, Hwann-Tzong Chen, and Tyng-Luh Liu.
\newblock Text-guided graph neural networks for referring 3d instance segmentation.
\newblock In \emph{AAAI}, 2021.

\bibitem[Hugo~Touvron et~al.(2023)Hugo~Touvron, Stone, Albert, Almahairi, Babaei, Bashlykov, Batra, Bhargava, Bhosale, and et~al]{llama2}
Louis~Martin Hugo~Touvron, Kevin Stone, Peter Albert, Amjad Almahairi, Yasmine Babaei, Nikolay Bashlykov, Soumya Batra, Prajjwal Bhargava, Shruti Bhosale, and et al.
\newblock Llama 2: Open foundation and fine-tuned chat models.
\newblock \emph{arXiv:2307.09288}, 2023.

\bibitem[Iyer et~al.(2022)Iyer, Lin, Pasunuru, Mihaylov, Simig, Yu, Shuster, Wang, Liu, Koura, Li, O'Horo, Pereyra, Wang, Dewan, Celikyilmaz, Zettlemoyer, and Stoyanov]{opt-iml}
Srinivasan Iyer, Xi~Victoria Lin, Ramakanth Pasunuru, Todor Mihaylov, Daniel Simig, Ping Yu, Kurt Shuster, Tianlu Wang, Qing Liu, Punit~Singh Koura, Xian Li, Brian O'Horo, Gabriel Pereyra, Jeff Wang, Christopher Dewan, Asli Celikyilmaz, Luke Zettlemoyer, and Ves Stoyanov.
\newblock Opt-iml: Scaling language model instruction meta learning through the lens of generalization.
\newblock \emph{arXiv preprint arXiv:2212.12017}, 2022.

\bibitem[Jin et~al.(2023)Jin, Hayat, Yang, Guo, and Lei]{jin2023context}
Zhao Jin, Munawar Hayat, Yuwei Yang, Yulan Guo, and Yinjie Lei.
\newblock Context-aware alignment and mutual masking for 3d-language pre-training.
\newblock In \emph{CVPR}, 2023.

\bibitem[Kolodiazhnyi et~al.(2023)Kolodiazhnyi, Vorontsova, Konushin, and Rukhovich]{kolodiazhnyi2023oneformer3d}
Maxim Kolodiazhnyi, Anna Vorontsova, Anton Konushin, and Danila Rukhovich.
\newblock Oneformer3d: One transformer for unified point cloud segmentation.
\newblock \emph{arXiv preprint arXiv:2311.14405}, 2023.

\bibitem[Lai et~al.(2024)Lai, Tian, Chen, Li, Yuan, Liu, and Jia]{lai2023lisa}
Xin Lai, Zhuotao Tian, Yukang Chen, Yanwei Li, Yuhui Yuan, Shu Liu, and Jiaya Jia.
\newblock Lisa: Reasoning segmentation via large language model.
\newblock In \emph{CVPR}, 2024.

\bibitem[Landrieu and Simonovski(2018)]{superpoint}
Loic Landrieu and Martin Simonovski.
\newblock Large-scale point cloud semantic segmentation with superpoint graphs.
\newblock In \emph{CVPR}, 2018.

\bibitem[Landrieu and Simonovsky(2018)]{Loic2018super}
Loic Landrieu and Martin Simonovsky.
\newblock Large-scale point cloud semantic segmentation with superpoint graphs.
\newblock In \emph{CVPR}, 2018.

\bibitem[Li et~al.(2023)Li, Li, Savarese, and Hoi]{li2023blip2}
Junnan Li, Dongxu Li, Silvio Savarese, and Steven Hoi.
\newblock Blip-2: Bootstrapping language-image pre-training with frozen image encoders and large language models.
\newblock \emph{arXiv preprint arXiv:2301.12597}, 2023.

\bibitem[Lin(2004)]{rouge}
Chin-Yew Lin.
\newblock Rouge: A package for automatic evaluation of summaries.
\newblock In \emph{Text summarization branches out}, 2004.

\bibitem[Liu et~al.(2023)Liu, Li, Wu, and Lee]{liu2023llava}
Haotian Liu, Chunyuan Li, Qingyang Wu, and Yong~Jae Lee.
\newblock Visual instruction tuning.
\newblock In \emph{NeurIPS}, 2023.

\bibitem[Luo et~al.(2022)Luo, Fu, Kong, Gao, Ren, Shen, Xia, and Liu]{luo20223d}
Junyu Luo, Jiahui Fu, Xianghao Kong, Chen Gao, Haibing Ren, Hao Shen, Huaxia Xia, and Si Liu.
\newblock 3d-sps: Single-stage 3d visual grounding via referred point progressive selection.
\newblock \emph{arXiv preprint arXiv:2204.06272}, 2022.

\bibitem[Nguyen et~al.(2024)Nguyen, Ngo, Kalogerakis, Gan, Tran, Pham, and Nguyen]{nguyen2023open3dis}
Phuc D.~A. Nguyen, Tuan~Duc Ngo, Evangelos Kalogerakis, Chuang Gan, Anh Tran, Cuong Pham, and Khoi Nguyen.
\newblock Open3dis: Open-vocabulary 3d instance segmentation with 2d mask guidance.
\newblock In \emph{CVPR}, 2024.

\bibitem[Papineni et~al.(2002)Papineni, Roukos, Ward, and Zhu]{bleu}
Kishore Papineni, Salim Roukos, Todd Ward, and Wei-Jing Zhu.
\newblock Bleu: a method for automatic evaluation of machine translation.
\newblock In \emph{ACL}, 2002.

\bibitem[Peng et~al.(2023)Peng, Genova, Jiang, Tagliasacchi, Pollefeys, and Funkhouser]{Peng2023OpenScene}
Songyou Peng, Kyle Genova, Chiyu~"Max" Jiang, Andrea Tagliasacchi, Marc Pollefeys, and Thomas Funkhouser.
\newblock Openscene: 3d scene understanding with open vocabularies.
\newblock In \emph{CVPR}, 2023.

\bibitem[Qi et~al.(2024{\natexlab{a}})Qi, Chen, Yang, Shen, Li, Guo, Xu, and Yang]{qi2024generalizable}
Lu Qi, Yi-Wen Chen, Lehan Yang, Tiancheng Shen, Xiangtai Li, Weidong Guo, Yu Xu, and Ming-Hsuan Yang.
\newblock Generalizable entity grounding via assistance of large language model.
\newblock \emph{arXiv preprint arXiv:2402.02555}, 2024{\natexlab{a}}.

\bibitem[Qi et~al.(2024{\natexlab{b}})Qi, Dong, Zhang, Geng, Han, Ge, Wang, Yi, and Ma]{qi2024shapellm}
Zekun Qi, Runpei Dong, Shaochen Zhang, Haoran Geng, Chunrui Han, Zheng Ge, He Wang, Li Yi, and Kaisheng Ma.
\newblock Shapellm: Universal 3d object understanding for embodied interaction.
\newblock \emph{arXiv preprint arXiv:2402.17766}, 2024{\natexlab{b}}.

\bibitem[Qi et~al.(2024{\natexlab{c}})Qi, Fang, Sun, Wu, Wu, Wang, Lin, and Zhao]{GPT4Point}
Zhangyang Qi, Ye Fang, Zeyi Sun, Xiaoyang Wu, Tong Wu, Jiaqi Wang, Dahua Lin, and Hengshuang Zhao.
\newblock Gpt4point: A unified framework for point-language understanding and generation.
\newblock In \emph{CVPR}, 2024{\natexlab{c}}.

\bibitem[Qian et~al.(2024)Qian, Ma, Ji, and Sun]{xrefseg3d}
Zhipeng Qian, Yiwei Ma, Jiayi Ji, and Xiaoshuai Sun.
\newblock X-refseg3d: Enhancing referring 3d instance segmentation via structured cross-modal graph neural networks.
\newblock In \emph{AAAI}, 2024.

\bibitem[Radford et~al.(2021)Radford, Kim, Hallacy, Ramesh, Goh, Agarwal, Sastry, Askell, Mishkin, Clark, Krueger, and Sutskever]{clip}
Alec Radford, Jong~Wook Kim, Chris Hallacy, Aditya Ramesh, Gabriel Goh, Sandhini Agarwal, Girish Sastry, Amanda Askell, Pamela Mishkin, Jack Clark, Gretchen Krueger, and Ilya Sutskever.
\newblock Learning transferable visual models from natural language supervision.
\newblock \emph{arXiv preprint arXiv:2103.00020}, 2021.

\bibitem[Rasheed et~al.(2024)Rasheed, Maaz, Shaji, Shaker, Khan, Cholakkal, Anwer, Xing, Yang, and Khan]{hanoona2023GLaMM}
Hanoona Rasheed, Muhammad Maaz, Sahal Shaji, Abdelrahman Shaker, Salman Khan, Hisham Cholakkal, Rao~M. Anwer, Eric Xing, Ming-Hsuan Yang, and Fahad~S. Khan.
\newblock Glamm: Pixel grounding large multimodal model.
\newblock In \emph{CVPR}, 2024.

\bibitem[Ren et~al.(2023)Ren, Huang, Wei, Zhao, Fu, Feng, and Jin]{ren2023pixellm}
Zhongwei Ren, Zhicheng Huang, Yunchao Wei, Yao Zhao, Dongmei Fu, Jiashi Feng, and Xiaojie Jin.
\newblock Pixellm: Pixel reasoning with large multimodal model.
\newblock \emph{arXiv preprint arXiv:2312.02228}, 2023.

\bibitem[Rozenberszki et~al.(2022)Rozenberszki, Litany, and Dai]{rozenberszki2022language}
David Rozenberszki, Or Litany, and Angela Dai.
\newblock Language-grounded indoor 3d semantic segmentation in the wild.
\newblock In \emph{ECCV}, 2022.

\bibitem[Schult et~al.(2023)Schult, Engelmann, Hermans, Litany, Tang, and Leibe]{Schult23ICRA}
Jonas Schult, Francis Engelmann, Alexander Hermans, Or Litany, Siyu Tang, and Bastian Leibe.
\newblock {Mask3D: Mask Transformer for 3D Semantic Instance Segmentation}.
\newblock In \emph{ICRA}, 2023.

\bibitem[Sun et~al.(2022)Sun, Qing, Tan, and Xu]{spformer}
Jiahao Sun, Chunmei Qing, Junpeng Tan, and Xiangmin Xu.
\newblock Superpoint transformer for 3d scene instance segmentation.
\newblock \emph{arXiv preprint arXiv:2211.15766}, 2022.

\bibitem[Takmaz et~al.(2023)Takmaz, Fedele, Sumner, Pollefeys, Tombari, and Engelmann]{takmaz2023openmask3d}
Ay{\c{c}}a Takmaz, Elisabetta Fedele, Robert~W. Sumner, Marc Pollefeys, Federico Tombari, and Francis Engelmann.
\newblock {OpenMask3D: Open-Vocabulary 3D Instance Segmentation}.
\newblock In \emph{NeurIPS}, 2023.

\bibitem[Tang et~al.(2024)Tang, Han, Li, Yu, Hao, Hu, and Chen]{tang2024minigpt_3d}
Yuan Tang, Xu Han, Xianzhi Li, Qiao Yu, Yixue Hao, Long Hu, and Min Chen.
\newblock Minigpt-3d: Efficiently aligning 3d point clouds with large language models using 2d priors.
\newblock \emph{arXiv preprint arXiv:2405.01413}, 2024.

\bibitem[teams(2023)]{Achiam2023GPT4TR}
OpenAI teams.
\newblock Gpt-4 technical report.
\newblock \emph{arXiv preprint arXiv:2303.08774}, 2023.

\bibitem[Touvron et~al.(2023)Touvron, Lavril, Izacard, Martinet, Lachaux, Lacroix, Rozière, Goyal, Hambro, Azhar, Rodriguez, Joulin, Grave, and Lample]{llama}
Hugo Touvron, Thibaut Lavril, Gautier Izacard, Xavier Martinet, Marie-Anne Lachaux, Timothée Lacroix, Baptiste Rozière, Naman Goyal, Eric Hambro, Faisal Azhar, Aurelien Rodriguez, Armand Joulin, Edouard Grave, and Guillaume Lample.
\newblock Llama: Open and efficient foundation language models.
\newblock \emph{arXiv:2302.13971}, 2023.

\bibitem[Vedantam et~al.(2015)Vedantam, Zitnick, and Parikh]{cider}
Ramakrishna Vedantam, C~Lawrence Zitnick, and Devi Parikh.
\newblock { Cider: Consensus-based image description evaluation}.
\newblock In \emph{CVPR}, 2015.

\bibitem[Wang(2023)]{Wang2023OctFormer}
Peng-Shuai Wang.
\newblock Octformer: Octree-based transformers for {3D} point clouds.
\newblock In \emph{SIGGRAPH}, 2023.

\bibitem[Wu et~al.(2024)Wu, Ma, Chen, Wang, Luo, Ji, and Sun]{wu2024stmn}
Changli Wu, Yiwei Ma, Qi Chen, Haowei Wang, Gen Luo, Jiayi Ji, and Xiaoshuai Sun.
\newblock 3d-stmn: Dependency-driven superpoint-text matching network for end-to-end 3d referring expression segmentation.
\newblock In \emph{AAAI}, 2024.

\bibitem[Wu and Xie(2023)]{vstar}
Penghao Wu and Saining Xie.
\newblock V*: Guided visual search as a core mechanism in multimodal llms.
\newblock \emph{arXiv preprint arXiv:2312.14135}, 2023.

\bibitem[Wu et~al.(2023)Wu, Biamby, Chan, Dunlap, Gupta, Wang, Gonzalez, and Darrell]{wu2023see}
Tsung-Han Wu, Giscard Biamby, David Chan, Lisa Dunlap, Ritwik Gupta, Xudong Wang, Joseph~E. Gonzalez, and Trevor Darrell.
\newblock See, say, and segment: Teaching lmms to overcome false premises.
\newblock \emph{arXiv preprint arXiv:2312.08366}, 2023.

\bibitem[Wu et~al.(2022)Wu, Lao, Jiang, Liu, and Zhao]{wu2022point}
Xiaoyang Wu, Yixing Lao, Li Jiang, Xihui Liu, and Hengshuang Zhao.
\newblock Point transformer v2: Grouped vector attention and partition-based pooling.
\newblock In \emph{NeurIPS}, 2022.

\bibitem[Xu et~al.(2023)Xu, Wang, Wang, Chen, Pang, and Lin]{xu2023pointllm}
Runsen Xu, Xiaolong Wang, Tai Wang, Yilun Chen, Jiangmiao Pang, and Dahua Lin.
\newblock Pointllm: Empowering large language models to understand point clouds.
\newblock \emph{arXiv preprint arXiv:2308.16911}, 2023.

\bibitem[Ye et~al.(2023)Ye, Xu, Xu, Ye, Yan, Zhou, Wang, Hu, Shi, Shi, Li, Xu, Chen, Tian, Qian, Zhang, Huang, and Zhou]{ye2023mplugowl}
Qinghao Ye, Haiyang Xu, Guohai Xu, Jiabo Ye, Ming Yan, Yiyang Zhou, Junyang Wang, Anwen Hu, Pengcheng Shi, Yaya Shi, Chenliang Li, Yuanhong Xu, Hehong Chen, Junfeng Tian, Qi Qian, Ji Zhang, Fei Huang, and Jingren Zhou.
\newblock mplug-owl: Modularization empowers large language models with multimodality.
\newblock \emph{arXiv preprint arXiv:2304.14178}, 2023.

\bibitem[Yuan et~al.(2021)Yuan, Yan, Liao, Zhang, Li, and Cui]{yuan2021instancerefer}
Zhihao Yuan, Xu Yan, Yinghong Liao, Ruimao Zhang, Zhen Li, and Shuguang Cui.
\newblock Instancerefer: Cooperative holistic understanding for visual grounding on point clouds through instance multi-level contextual referring.
\newblock In \emph{ICCV}, 2021.

\bibitem[Zhan et~al.(2024)Zhan, Dai, Ye, Zhou, Zhang, Liu, Zhang, Yuan, Zhang, Li, et~al.]{zhan2024anygpt}
Jun Zhan, Junqi Dai, Jiasheng Ye, Yunhua Zhou, Dong Zhang, Zhigeng Liu, Xin Zhang, Ruibin Yuan, Ge Zhang, Linyang Li, et~al.
\newblock Anygpt: Unified multimodal llm with discrete sequence modeling.
\newblock \emph{arXiv preprint arXiv:2402.12226}, 2024.

\bibitem[Zhang et~al.(2023{\natexlab{a}})Zhang, Sun, Chen, Xiao, Shao, Zhang, Chen, and Luo]{zhang2023gpt4roi}
Shilong Zhang, Peize Sun, Shoufa Chen, Min Xiao, Wenqi Shao, Wenwei Zhang, Kai Chen, and Ping Luo.
\newblock Gpt4roi: Instruction tuning large language model on region-of-interest.
\newblock \emph{arXiv preprint arXiv:2307.03601}, 2023{\natexlab{a}}.

\bibitem[Zhang et~al.(2023{\natexlab{b}})Zhang, Zhang, Li, Zhao, Karypis, and Smola]{zhang2023multicot}
Zhuosheng Zhang, Aston Zhang, Mu Li, Hai Zhao, George Karypis, and Alex Smola.
\newblock Multimodal chain-of-thought reasoning in language models.
\newblock \emph{arXiv preprint arXiv:2302.00923}, 2023{\natexlab{b}}.

\bibitem[Zhao et~al.(2021{\natexlab{a}})Zhao, Jiang, Jia, Torr, and Koltun]{zhao2021point}
Hengshuang Zhao, Li Jiang, Jiaya Jia, Philip~HS Torr, and Vladlen Koltun.
\newblock Point transformer.
\newblock In \emph{ICCV}, 2021{\natexlab{a}}.

\bibitem[Zhao et~al.(2021{\natexlab{b}})Zhao, Cai, Sheng, and Xu]{zhao2021_3DVG_Transformer}
Lichen Zhao, Daigang Cai, Lu Sheng, and Dong Xu.
\newblock {3DVG-Transformer}: Relation modeling for visual grounding on point clouds.
\newblock In \emph{ICCV}, 2021{\natexlab{b}}.

\bibitem[Zheng et~al.(2023)Zheng, Chiang, Sheng, Zhuang, Wu, Zhuang, Lin, Li, Li, Xing, Zhang, Gonzalez, and Stoica]{zheng2023judging}
Lianmin Zheng, Wei-Lin Chiang, Ying Sheng, Siyuan Zhuang, Zhanghao Wu, Yonghao Zhuang, Zi Lin, Zhuohan Li, Dacheng Li, Eric.~P Xing, Hao Zhang, Joseph~E. Gonzalez, and Ion Stoica.
\newblock Judging llm-as-a-judge with mt-bench and chatbot arena.
\newblock \emph{arXiv preprint arXiv:2306.05685}, 2023.

\bibitem[Zhu et~al.(2023)Zhu, Chen, Shen, Li, and Elhoseiny]{zhu2023minigpt}
Deyao Zhu, Jun Chen, Xiaoqian Shen, Xiang Li, and Mohamed Elhoseiny.
\newblock Minigpt-4: Enhancing vision-language understanding with advanced large language models.
\newblock \emph{arXiv preprint arXiv:2304.10592}, 2023.

\end{thebibliography}
}

\maketitlesupplementary
\renewcommand\thesection{\Alph{section}}
\setcounter{section}{0}
We provide additional details on our dataset in Section~\ref{sec:supp_dataset}. Section~\ref{sec:supp_exp} elaborates on the implementation and presents further results. Sections~\ref{sec:supp_failure} to~\ref{sec:impact} offer comprehensive discussions on potential failure cases, limitations, and the broader impact of our method.

\section{Reason3D Dataset}
\label{sec:supp_dataset}
\subsection{Dataset Annotation} 
Each 3D scene in the Reason3D dataset consists of a textual query and a binary segmentation mask to identify the target objects. As mentioned in the paper, we utilize ScanNetV2~\cite{dai2017scannet, rozenberszki2022language} and Matterport3D~\cite{Matterport3D} as our data sources. Considering single room space as one scene, we first extract the instance object annotation and the room type information from these two datasets as the tags of 3D scenes. After that, we utilize these tags as parts of the text prompt to incorporate with GPT-4. 
The illustration of the prompt construction process is shown in Table~\ref{tab:prompt_conversation}, and some samples utilized for prompting are shown in Table~\ref{tab:few}.

\begin{table*}[ht]\centering
\begin{minipage}{0.99\textwidth}\vspace{0mm}    \centering
\begin{tcolorbox} 
    \centering
    \small
     \hspace{-6mm}
    \begin{tabular}{p{0.99\columnwidth}}

\begin{minipage}{0.99\columnwidth}\vspace{0mm}

\VarSty{messages} = [
            \{\var{"role":"system", "content":} You are an AI visual assistant, and you are seeing a 3D scene. 
            What you see is provided with several words to represent objects with tag  {\bf <objects>}, describing the scene you are looking at, and also the room type {\bf <type>} to describe the type of the scene. 
            Design a question {\bf <question>} that can be answered confidently with the {\bf <answer>} from one of the provided objects in {\bf <objects>}. 
            Please do not ask any {\bf <question>} that cannot be answered confidently. 
            Each question should have one clear answer that is most relevant, without ambiguity or multiple possible answers in the list of description words.  
            Please include complex questions relevant to the scene's content, such as inquiries into the background knowledge of objects or discussions about events related to these objects. Avoid questions about uncertain or unclear details. 
            The question should be natural.\}\\]
    
  \For{ \VarSty{sample} in   \VarSty{few\_shot\_samples}}{
         \var{\VarSty{messages}.append(\{"role":"user", "content":\VarSty{sample[`context']}\})} \; \\
         \var{\VarSty{messages}.append(\{"role":"assistant", "content":\VarSty{sample[`response']}\} ) } \;
         }  
    \var{\VarSty{messages}.append(\{"role":"user", "content":`\textbackslash  n'.join(\VarSty{query})\})}
\end{minipage}
\end{tabular}
\end{tcolorbox}
    
\vspace{-2mm}
\caption{The illustration of the prompt construction process for generating 3D reasoning dataset with ChatGPT / GPT-4.}
\label{tab:prompt_conversation}
\end{minipage}
\end{table*}

\begin{table*}[h!]\centering

\begin{minipage}{0.99\textwidth}\vspace{0mm}    \centering
\begin{tcolorbox} 
    \centering
    \small
     \hspace{-6mm}
    \begin{tabular}{p{0.99\columnwidth}}

\begin{minipage}{0.99\columnwidth}\vspace{0mm}
\var{\VarSty{`context'}}: {\bf<room>} game room {\bf<objects>} 'armchair', 'ceiling',  'door', 'doorframe', 'fireplace', 'floor', 'pool table',
 'post', 'rail', 'stair', 'stool', 'tv', 'wall', 'window' \; \\
\var{\VarSty{`response'}}: {\bf<question>} In a game room, what object in the scene could be used for playing a competitive and strategic game involving balls and cues? {\bf<answer>} pool table  \; 
\end{minipage}
    \end{tabular}
\end{tcolorbox}

\begin{tcolorbox} 
    \centering
    \small
     \hspace{-6mm}
    \begin{tabular}{p{0.99\columnwidth}}
\begin{minipage}{0.99\columnwidth}\vspace{0mm}
\var{\VarSty{`context'}}: {\bf<room>} living room {\bf<objects>} 'armchair', 'bedroom', 'bookshelf', 'lamp', 'bureau', 'carpet', 'ceiling fan', 'chair', 'computer', 'computer desk', 'couch', 'table', 'drawer', 'dresser' \; \\
\var{\VarSty{`response'}}: {\bf<question>} It’s very hot outside. After coming back home, what appliance would you turn on to help cool down the temperature? {\bf<answer>} ceiling fan  \; 
\end{minipage}
    \end{tabular}
\end{tcolorbox}

\begin{tcolorbox} 
    \centering
    \small
     \hspace{-6mm}
    \begin{tabular}{p{0.99\columnwidth}}
\begin{minipage}{0.99\columnwidth}\vspace{0mm}
\var{\VarSty{`context'}}: {\bf<room>} game room {\bf<objects>} 'table', 'door', 'cabinet', 'desk', 'office chair', 'picture', 'lamp', 'bathtub', 'bag', 'trash can', 'mirror', 'radiator' \; \\
\var{\VarSty{`response'}}: {\bf<question>} When staying at a hotel, what part of the room in the scene can provide additional lighting for reading or working while in bed? {\bf<answer>} lamp \; 
\end{minipage}
    \end{tabular}
\end{tcolorbox}

\begin{tcolorbox} 
    \centering
    \small
     \hspace{-6mm}
    \begin{tabular}{p{0.99\columnwidth}}
\begin{minipage}{0.99\columnwidth}\vspace{0mm}
\var{\VarSty{`context'}}: {\bf<room>} game room {\bf<objects>} 'floor', 'door', 'cabinet', 'shelf', 'desk', 'office chair', 'window', 'monitor', 'book', 'box', 'keyboard', 'trash can', 'file cabinet', 'fan', 'telephone', 'cup', 'paper towel roll', 'windowsill', 'clock' , 'headphones' \; \\
\var{\VarSty{`response'}}: {\bf<question>} If someone wanted to check the time after getting ready in the morning, what object in this scene would they most likely use? {\bf<answer>} clock \; 
\end{minipage}
    \end{tabular}
\end{tcolorbox}

\vspace{-2mm}
\caption{The few shot samples used for ChatGPT prompting.}
\label{tab:few}
\end{minipage}
\end{table*}

\subsection{Dataset Statistics} 
The collected Reason3D dataset incorporates the ScanNetV2~\cite{dai2017scannet} and Matterport3D~\cite{Matterport3D} datasets. We adhere to their official training and validation splits for data annotation. Specifically, the Matterport3D dataset provides 934 training samples and 837 validation samples. Meanwhile, the ScanNetV2 dataset contributes 405 training samples and 308 validation samples.

\subsection{3D Hierarchical Searching Extension} 
The 3D hierarchical searching task extends the reasoning-based 3D segmentation task by incorporating a specified target room type where the queried object should be located.
As detailed in the main paper, we utilize only a subset of the Matterport3D dataset, chosen for its diversity in room types. In this task, we include the template "In <ROOM\_TYPE>" to specify the target room, where <ROOM\_TYPE> represents the various room categories defined in the Matterport3D dataset.
For experiments involving different numbers of rooms, we expand the target room's space by including its neighboring rooms, adjusting this space according to the specified number of rooms.

\section{Experiments}
\label{sec:supp_exp}

\subsection{Implementation Details}
Our models are primarily executed on two NVIDIA RTX A6000 GPUs, using a batch size of 16 during training and 1 during inference.
The AdamW optimizer is employed with parameters $\beta_1=0.9$ and $\beta_2=0.999$, and a weight decay of 0.05. Additionally, we implement a linear warm-up strategy for the learning rate during the initial 1,000 steps, gradually increasing it from $10^{-8}$ to $10^{-4}$, followed by a cosine decay schedule.
All experiments are conducted using the PyTorch framework.

\subsection{Used Datasets}
In addition to the reason3D dataset, our model utilizes the following datasets for training:

\smallskip{\noindent{\bf ScanNet~\cite{dai2017scannet, rozenberszki2022language}}}, a comprehensive 3D indoor dataset, covers diverse environments including apartments and various room types. The dataset is structured into 1201 training scenes, 312 validation scenes, and 100 testing scenes. 

\smallskip{\noindent{\bf Matterpor3D~\cite{Matterport3D}}} dataset is a large-scale, real-world dataset comprising 90 houses. Each house is divided into various regions. We do not fine-tune the scene encoder on the Matterport3D dataset.

\smallskip{\noindent{\bf ScanRefer~\cite{chen2020scanrefer}}}, a dataset annotated using ScanNet for 3D express referring segmentation tasks, including 36,665 natural language descriptions related to 7,875 objects across 562 scenes for training, and 9,508 descriptions of 2,068 objects from 141 ScanNet scenes for evaluation.

\smallskip{\noindent{\bf Nr3D~\cite{achlioptas2020referit_3d}}}, another 3D referring segmentation dataset derived from ScanNet, comprises 32,919 language descriptions associated with 4,664 objects from 511 scenes for training purposes. We further employ this dataset to train for 3D express referring segmentation tasks.

\smallskip{\noindent{\bf ScanQA~\cite{azuma_2022_CVPR}}} is a dataset for the 3D question answering task based on ScanNet, consisting of 25,563 question-answer pairs on 562 scenes for training and  4,675 question-answer pairs on 71 scenes for validation.

\subsection{Training}
Following the approach in~\cite{spformer, wu2024stmn}, we first train a sparse 3D UNet~\cite{3dsparse} as our backbone and compute superpoint features using pre-computed superpoints~\cite{super, Loic2018super}. Once the backbone is fully trained, its weights are frozen. For each target task, we then pre-train the entire network using data from the other tasks, followed by task-specific fine-tuning.

\subsection{Instruction Template} 
In this section, we present the instructions and outputs used for task-specific templates. Following previous works~\cite{xu2023pointllm,chen2023ll3da}, we utilize a “human:“ identifier to initiate the instruction, followed by an “assistant:“ identifier for the LLM-generated response.
We use \textcolor{plotbrown}{<scene>} to represent the token corresponding to the point cloud scene. The tokens \textcolor{plotred}{<SEG>} and \textcolor{plotblue}{<LOC>} are used for segmentation and location, respectively, as part of the prompting process for generating segmentation results, as described in the main paper.
Below are examples for various tasks, where \{description\} refers to the target object's description, \{question\} represents a query based on the given scene, and \{answer\} is the corresponding response.

\begin{table*}[h]
\begin{minipage}[h]{.52\linewidth}
\centering
\vspace{1mm}
\footnotesize
\setlength\tabcolsep{3.7pt}
\begin{tabular}{lccccc}
\toprule
Method & Venue & B-4 & METEOR & ROUHE-L & CIDER \\
\midrule
VoteNet+MCAN & - & 6.2 & 11.4 & 29.8 & 54.7 \\
ScanRefer+MCAN & - & 7.9 & 11.5 & 30 & 55.4 \\
ScanQA~\cite{azuma_2022_CVPR} & CVPR 2022 &  10.1 & 13.1 & 33.3 & 64.9 \\
3D-VLP~\cite{jin2023context} & CVPR 2023 & 11.2 & 13.5 & 34.5 & 67.0\\
3D-LLM~\cite{3dllm} & NeurIPS 2023 & 12.0 & 14.5 & 35.7 & 69.4 \\
\midrule
Reason3D (Ours) & - & \textbf{12.1} & \textbf{15.1} & \textbf{37.4} & 
\textbf{73.5}  \\
\bottomrule
\end{tabular}
\caption{\textbf{3D question answering results} on ScanQA validation dataset. The first two results are from~\cite{azuma_2022_CVPR}. B-4 denotes BLEU-4. Our model achieves better results than all baseline models.}
\label{tab:3dqa}
\end{minipage}\hfill
\begin{minipage}[h]{.45\linewidth}
\centering
\vspace{1mm}
\footnotesize
\setlength\tabcolsep{5pt}
\begin{tabular}{lccc}
\toprule
Method & Venue & Acc@0.25 & Acc@0.50  \\
\midrule
ScanRefer~\cite{chen2020scanrefer} & ECCV 2020 & 38.97 & 26.10\\
InstanceRefer~\cite{yuan2021instancerefer} & ICCV 2021 & 40.23 & 32.93\\
3D-SPS~\cite{luo20223d} & CVPR 2022 &  47.65 & 36.43 \\
ViL3DRel~\cite{chen2022vil3dref} & NeurIPS 2022  & 47.94 & 37.73 \\
3D-LLM~\cite{3dllm} & NeurIPS 2023 & 30.3 & - \\ \midrule
TGNN~\cite{huang2021tgnn} & AAAI 2021  & 37.37 & 29.70  \\
3D-STMN~\cite{wu2024stmn} & AAAI 2024 &  46.8 & 36.6\\
\midrule
{Reason3D (Ours)} & - & \textbf{49.60} & \textbf{41.10}\\

\bottomrule
\end{tabular}
\caption{\textbf{3D visual grounding results} on ScanRefer validation dataset. Our approach does not use 3D box annotation for training.} 
\label{tab:3dvg}
\end{minipage}
\end{table*}

\smallskip{\noindent{\bf 3D Reasoning Segmentation:}}
“Human: \textcolor{plotbrown}{<scene>} can you segment the object in the scene with the following descriptions: {description}? Assistant: Sure, it's \textcolor{plotred}{<SEG>}.”

\smallskip{\noindent{\bf 3D Hierarchical Searching.}} 
“Human: \textcolor{plotbrown}{<scene>} can you segment the object based on the description: {description}? Please segment the target room first, then output the object mask. Assistant: Sure, the room is \textcolor{plotblue}{<LOC>}, and the object is \textcolor{plotred}{<SEG>}.”

\smallskip{\noindent{\bf 3D Express Referring Segmentation.}} 
“Human: \textcolor{plotbrown}{<scene>} please segment the object from the given scene: {description}. Assistant: It's \textcolor{plotred}{<SEG>}."

\smallskip{\noindent{\bf 3D Question Answering.}}
“Human: \textcolor{plotbrown}{<scene>} please answer the question based on the given scene: {question} and output the related segmentation mask. Assistant: {answer} \textcolor{plotred}{<SEG>}."

\subsection{ 3D Question Answering Results} 
\smallskip{\noindent{\bf Evaluation Metrics.}} For the QA task, the evaluation metrics include BLEU-4~\cite{bleu}, ROUGE-L~\cite{rouge}, METEOR~\cite{banarjee2005}, and CIDEr~\cite{cider} to ensure robust answer matching. These metrics evaluate the precision, fluency, and semantic accuracy of the responses.

\smallskip{\noindent{\bf Results.}} In addition to excelling in 3D reasoning and referring tasks, our approach also performs well in 3D question answering tasks. We present our results on the ScanQA validation set in Table~\ref{tab:3dqa}, where we observed a significant improvement in evaluation metrics over both baseline methods and the recent LLM-based method, 3D-LLM~\cite{3dllm}. Our approach not only answers questions accurately but also visualizes the related segmentation masks to further demonstrate the effectiveness. 

\subsection{3D Visual Grounding Results.} 
\smallskip{\noindent{\bf Evaluation Metrics.}}
Similar to the 3D expressive referring task, the evaluation metrics used are Acc@0.25 and Acc@0.5, indicating the percentage of correctly predicted bounding boxes with an IoU greater than 0.25 or 0.5, respectively, compared to the ground truth.

\smallskip{\noindent{\bf Results.}}
Although our primary focus is on 3D segmentation, our method also effectively predicts 3D bounding boxes as supplementary outputs, facilitating comparison with 3D visual grounding methods. To generate a 3D bounding box for a referred object, we first apply DBSCAN~\cite{dbscan} to eliminate noisy points, and then calculate the minimum and maximum XYZ coordinates from the points within the segmentation mask to form the 3D box. 
As demonstrated in Table~\ref{tab:3dvg}, our approach not only outperforms recent 3D visual grounding methods but also significantly surpasses LLM-based methods, such as 3D-LLM~\cite{3dllm}, which struggle to integrate textual and numerical data to accurately localize objects in 3D space.

\subsection{More Visualization 
Results.}
\begin{figure*}[t]
    \centering
\includegraphics[width=0.95\linewidth]{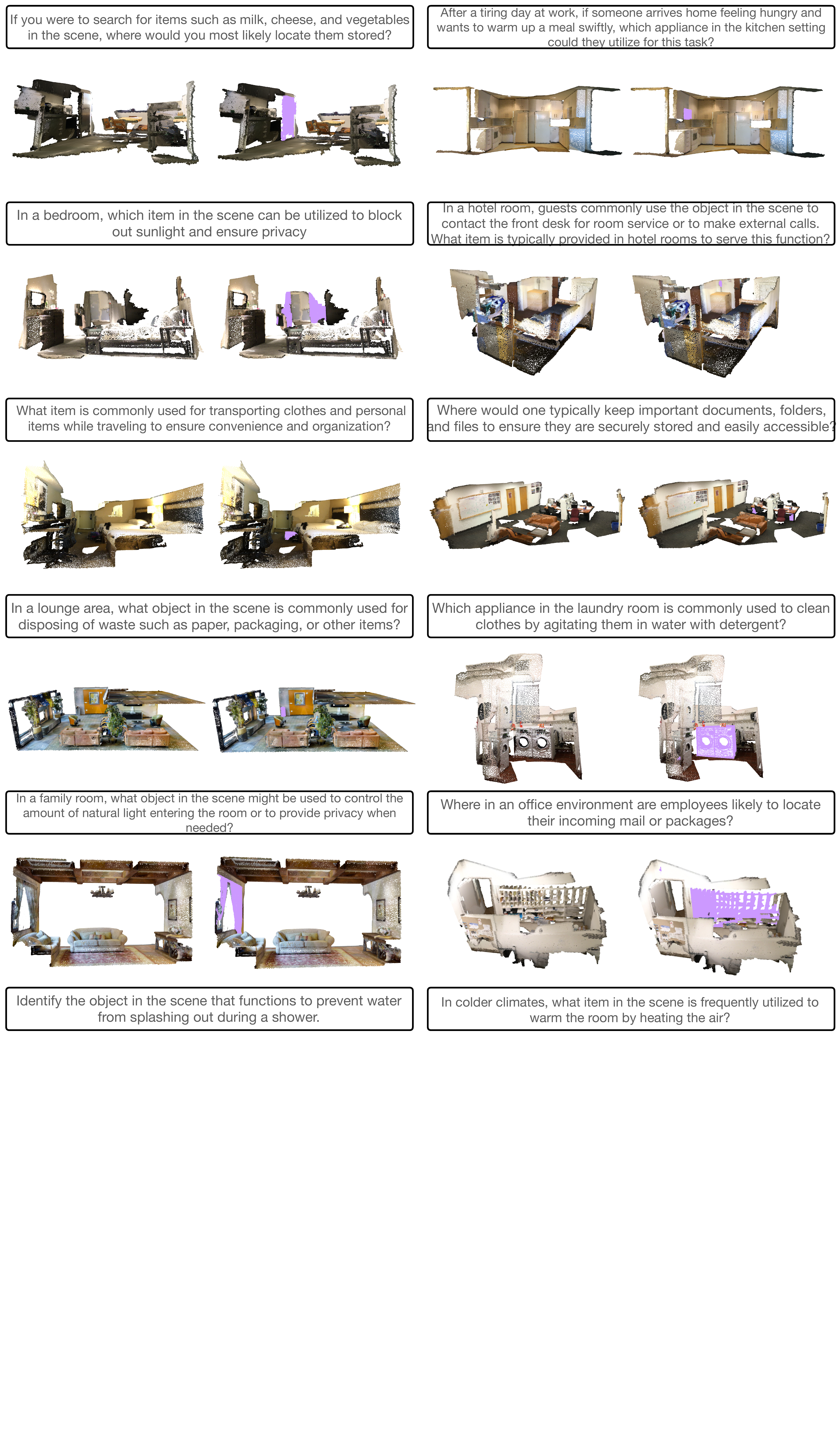}
    \vspace{-7.7cm}
    \caption{\textbf{Visualization Results for 3D Reasoning Segmentation Tasks.} The \textcolor{plotsamplepurple}{purple} regions highlight the predicted segmentation masks generated by our model. Best viewed with zoom in.}
    \label{fig:sup_reason}
\end{figure*}

\begin{figure*}[t]
    \centering
    \vspace{-0.8cm}
\includegraphics[width=0.85\linewidth]{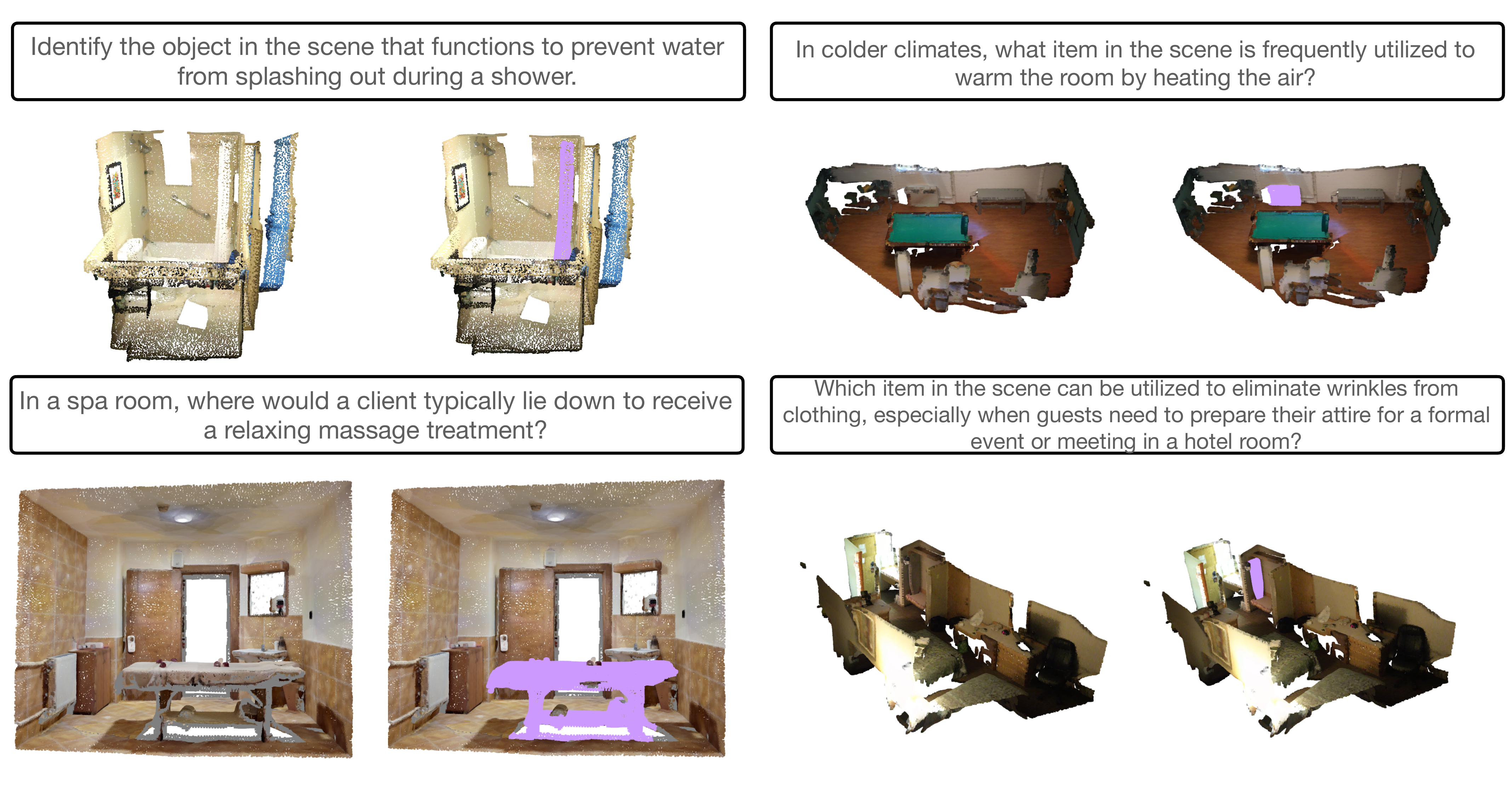}
    \vspace{-0.5cm}
    \caption{\textbf{More Visualization Results for 3D Reasoning Segmentation Tasks.} The \textcolor{plotsamplepurple}{purple} regions highlight the predicted segmentation masks generated by our model. Best viewed with zoom in.}
    \label{fig:sup_reason_2}
\end{figure*}

\noindent{\textbf{3D Reasoning Segmentation.}} 
We provide more qualitative examples for the 3D reasoning segmentation task and the predictions by our Reason3D in Figures~\ref{fig:sup_reason} and~\ref{fig:sup_reason_2}.

\begin{figure*}[t]
    \centering
\includegraphics[width=0.9\linewidth]{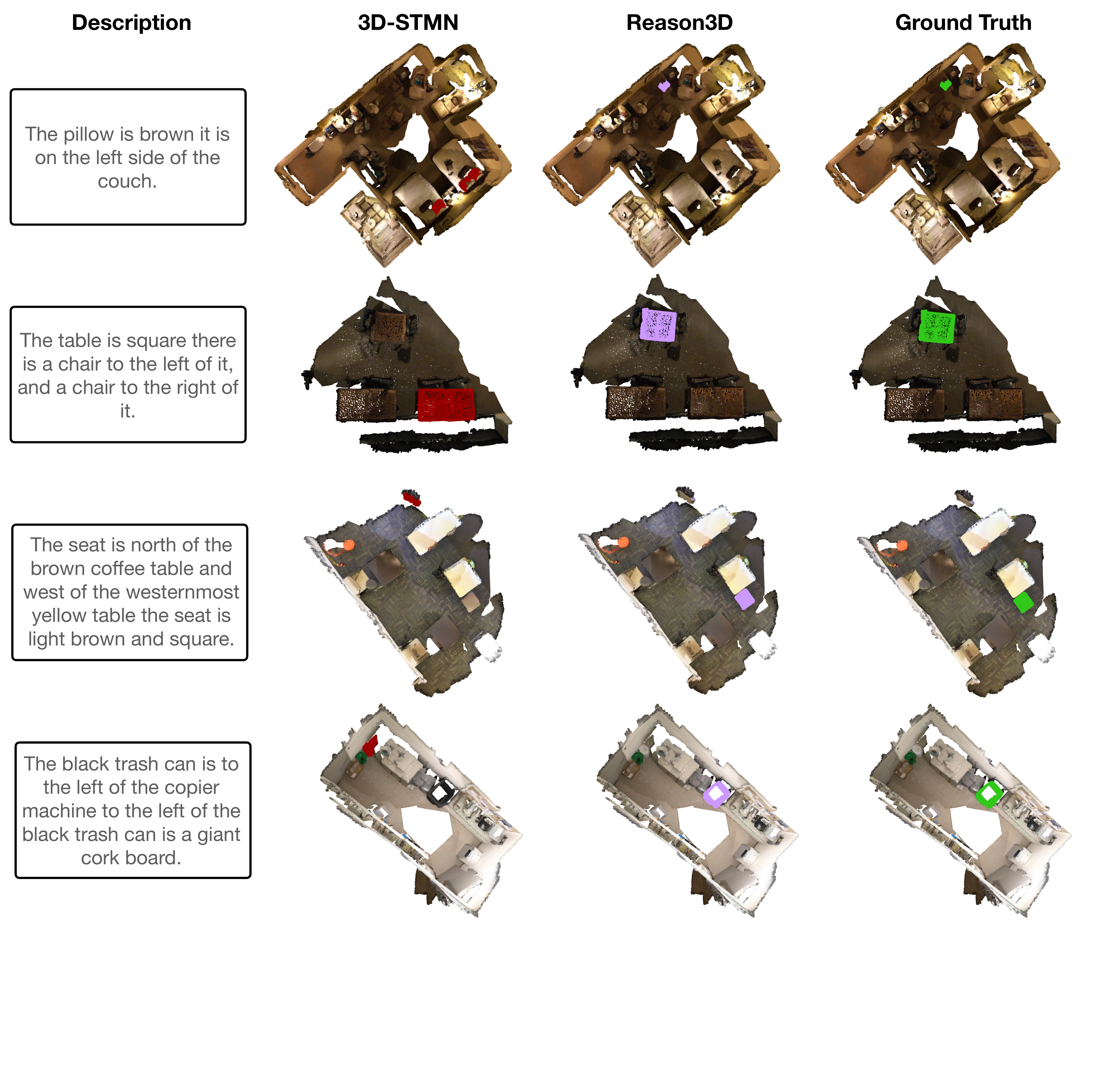}
    \vspace{-2.5cm}
    \caption{\textbf{Visualization Results for 3D Referring Segmentation Tasks.} The \textcolor{plotsamplepurple}{purple} regions denotes the predicted segmentation masks from our Reason3D. The \textcolor{plotsamplered}{red} and \textcolor{plotsamplegreen}{green} means the predictions from 3D-STMN and ground truth, respectively.   
    Best viewed with zoom in.}
    \label{fig:sup_refer}
\end{figure*}

\smallskip{\noindent{\textbf{3D Referring Segmentation.}}}
We show the visualization results of the 3D referring segmentation task compared with 3D-STMN~\cite{wu2024stmn} in Figure~\ref{fig:sup_refer}. We observe that our approach can have correct predictions when the scenes contain multiple similar objects or when the query sentence is long, which proves the effectiveness of our approach.

\section{Failure Case.}
\label{sec:supp_failure}
\begin{figure*}[h]
    \centering
\includegraphics[width=0.97\linewidth]{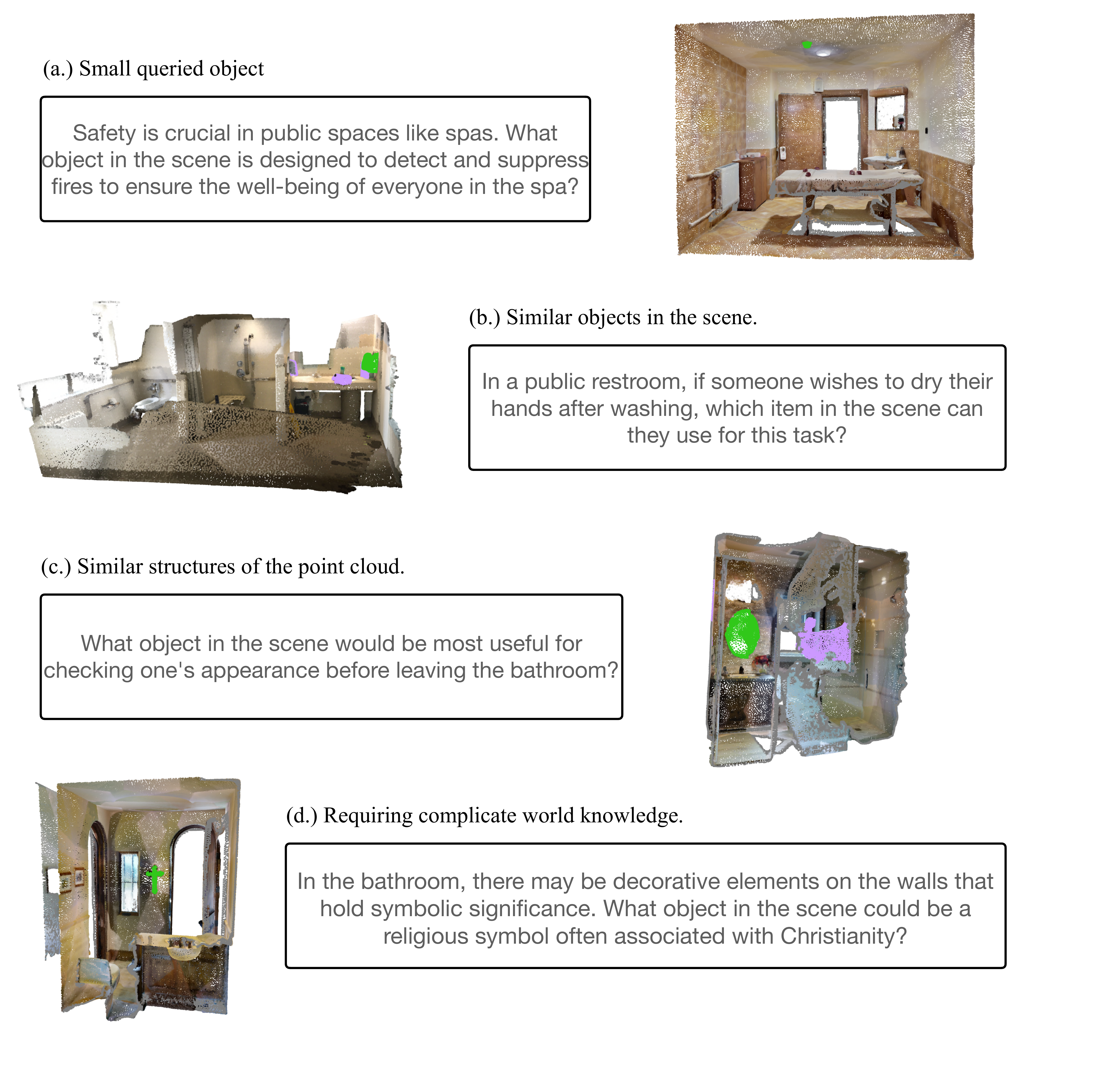}
    \vspace{-1.2cm}
    \caption{\textbf{Failure cases.} (a.) Small queried objects. (b.) Similar object in the scene. (c.) Similar structures of the point cloud. (d.) The question requires complicated world knowledge.
    The \textcolor{plotsamplepurple}{purple} regions denote the predicted segmentation masks from our Reason3D, and the \textcolor{plotsamplegreen}{green} means the ground truth.   
    Best viewed with zoom in.}
    \label{fig:sup_fail}
\end{figure*}

In Figure~\ref{fig:sup_fail}, we present representative failure cases as follows: (a) If the question involves querying a small object in the scene, our model may fail to generate the correct prediction. (b) The presence of similar objects in the scene may lead to false positive predictions by our model. (c) Similar structures in the point cloud, such as mirrors or sensor-induced fragments, can mislead our model. (d) Complex world knowledge required by the question may hinder our model's ability to generate accurate mask predictions.

\section{Limitations}
\label{sec:limitation}
While our model introduces a novel approach to 3D reasoning segmentation, it does have limitations that open promising avenues for future research. For example, our system currently struggles with large-scale scenes—such as identifying an object within a 30-room house in the Matterport dataset. Future work could refine the hierarchical decoder or integrate adaptive multi-scale processing techniques to better manage complex spatial extents.
Additionally, the model does not yet handle scenarios with false premises—such as querying for an object that may not be present—suggesting that incorporating uncertainty estimation or robust error-detection mechanisms could help validate query assumptions before processing. Moreover, since our implementation is designed primarily for single-object queries, its performance on multi-object or multi-category tasks remains untested. This limitation points to the need for developing enriched query representations and joint optimization strategies that can simultaneously manage multiple objects.
Addressing these challenges would significantly enhance the robustness and versatility of our LLM-based 3D reasoning framework.

\section{Broader Impact}
\label{sec:impact}
Reason3D is designed to segment objects in 3D space based on language inputs. Compared to traditional 3D segmentation algorithms, Reason3D models have a lower barrier to customization, enabling users to identify objects using natural language. However, this increased accessibility also raises the potential for misuse. Furthermore, the datasets and pre-trained models used in Reason3D may carry inherent biases, which could influence the model's performance.

\end{document}